\documentclass[conference, 10pt]{IEEEtran}
\usepackage{epsfig,endnotes}
\usepackage{url}
\usepackage[hidelinks]{hyperref}
\usepackage{xcolor,colortbl}
\usepackage{booktabs,caption,fixltx2e}
\usepackage[flushleft]{threeparttable}
\usepackage{wrapfig}
\usepackage{makecell}
\usepackage{float,graphicx}
\usepackage[font=small]{caption}
\usepackage{subcaption}
\usepackage{multicol}
\usepackage{multirow}
\usepackage{float}
\usepackage{tikz}
\usepackage{mwe}
\usepackage{listings}
\usepackage{amsmath,mathtools,amssymb}
\usepackage{array, makecell}
\usepackage{enumitem}
\usepackage{microtype}
\usepackage{verbatim}
\usepackage{setspace}
\usepackage[ruled,linesnumbered]{algorithm2e}

\SetCommentSty{mycommfont}
\usepackage{xcolor,colortbl}
\definecolor{ashgrey}{rgb}{0.7, 0.75, 0.71}
\SetKw{KwBy}{by}
\usepackage{textcomp}
\makeatletter
\g@addto@macro{\UrlBreaks}{\UrlOrds}

\definecolor{Gray}{gray}{0.85}
\definecolor{lblue}{rgb}{0.7,1,1}
\definecolor{codegray}{rgb}{0.5,0.5,0.5}
\definecolor{codepurple}{rgb}{0.58,0,0.82}
\definecolor{backcolour}{rgb}{0.95,0.95,0.92}

\usepackage{pifont}
\newcommand{\cmark}{\ding{51}}
\newcommand{\xmark}{\ding{55}}

\lstdefinestyle{mystyle}{
    backgroundcolor=\color{backcolour},   
    commentstyle=\color{codegray},
    keywordstyle=\color{magenta},
    numberstyle=\tiny\color{codegray},
    stringstyle=\color{codepurple},
    basicstyle=\ttfamily\scriptsize,
    breakatwhitespace=false,         
    breaklines=true,                 
    captionpos=b,                    
    keepspaces=true,                               
    numbersep=5pt,                  
    showspaces=false,                
    showstringspaces=false,
    showtabs=false,                  
    tabsize=2
}

\def\endthebibliography{%
  \def\@noitemerr{\@latex@warning{Empty `thebibliography' environment}}%
  \endlist
}
\ifCLASSINFOpdf
\else
\fi

\begin{document}

\title{ReDIL-GNN: Resynthesis Domain Incremental Learning for Circuit Graph Neural Networks}

\author{
  \IEEEauthorblockN{Rupesh Raj Karn, Johann Knechtel, Ozgur Sinanoglu}\vspace{-10pt}\\
  \IEEEauthorblockA{Center for Cyber Security, New York University, Abu Dhabi, UAE.}\vspace{-10pt}\\
  Email: \{rupesh.k, johann, ozgursin\}@nyu.edu
}


\maketitle

\begin{abstract}
Logic resynthesis preserves circuit functionality while changing gate vocabulary, topology, and structural statistics, creating domain shift for circuit graph neural networks (GNNs) without changing task labels. To study this setting, we introduce ReDIL-GNN, a resynthesis domain-incremental learning framework that adapts a fixed prediction or representation head as new synthesis styles arrive and evaluates retention on all previously observed domains. Because not every shift should be adapted blindly, ReDIL-GNN further introduces the Resynthesis Adaptability Index (RAI), a pre-adaptation score that combines adaptation need, source-equivalence recoverability, structural coverage, and update compatibility. We evaluate supervised hardware-security tasks and representation-learning models using task-native metrics for classifiers and source-equivalence retrieval metrics for embedding models, comparing naive fine-tuning with LwF, Online EWC, MAS, ER, A-GEM, DER++, ER+LwF, and equivalence-guided replay. Across the studied pipelines, RAI separates unsupported shifts from promising updates, ranging from \(0.001\) for a structurally uncovered GNN-RE \texttt{ABC-rewrite} shift to \(0.824\) for the best original-only GNN-RE adaptation case. In practice, ReDIL-GNN turns resynthesis-aware circuit learning into a deployment control loop: RAI screens each new synthesis flow before update, guiding whether to reuse the current model, apply retention-aware adaptation, or defer adaptation until the shift is better supported.
\end{abstract}

\IEEEpeerreviewmaketitle


\section{Introduction}
\label{sec:introduction}

A single Boolean functionality can be implemented by many structurally different gate-level netlists because logic rewriting, technology mapping, and LUT mapping can change the implementation while preserving the intended logic behavior~\cite{brayton2010abc,wolf2013yosys}. Circuit graph neural networks (GNNs) consume such structural representations and have been adopted for circuit-design \cite{zhang2019circuit}, reverse-engineering \cite{alrahis2021gnn}, hardware-security \cite{lashen13}, and circuit-representation tasks~\cite{yang2022versatile}. This creates a practical deployment mismatch: the circuit label or source identity may remain unchanged, while the graph distribution shifts through changed gate vocabularies, node counts, fan-in/fan-out patterns, logic depth, and connectivity~\cite{brayton2010abc,wolf2013yosys,amaru2015epfl}. Standard evaluation on a fixed synthesis distribution is therefore insufficient for measuring whether a circuit GNN can learn a newly observed synthesis style while retaining performance on previously observed styles~\cite{wang2024comprehensive,zhou2021overcoming,liu2021overcoming}. Moreover, not every new resynthesis style should trigger adaptation: some shifts are already handled by the base model, while others may be structurally unsupported or may induce forgetting if updated naively.


We study this setting as \emph{resynthesis-domain incremental learning (Domain-IL)}, where each incoming task is a synthesis domain and the semantic label space remains fixed, making the problem a domain-incremental rather than class-incremental continual-learning problem~\cite{wang2024comprehensive,wang2020streaming}. Unlike prior circuit continual-learning studies that evaluate task-incremental learning with artificial task splits~\cite{karn2026benchmarking,karn2026dynamic} or class-incremental learning with expanding gate vocabularies~\cite{karn2026cilcircuits}, we address \emph{Domain-IL}, where task labels and architectural heads remain invariant, but the graph topology undergoes semantics-preserving distribution shift induced by logic resynthesis. ReDIL-GNN starts from a model trained on original and previously observed resynthesized netlists, then adapts the same prediction or representation head as new synthesis styles arrive~\cite{li2016learning,kirkpatrick2017overcoming,aljundi2018memory}. We compare naive sequential fine-tuning with distillation-based~\cite{li2016learning}, regularization-based~\cite{kirkpatrick2017overcoming,aljundi2018memory}, replay-based~\cite{rolnick2019experience,chaudhry2019tiny}, gradient-projection-based~\cite{chaudhry2019efficient}, dark-replay~\cite{buzzega2020dark}, hybrid replay--distillation~\cite{rolnick2019experience,li2016learning}, and our equivalence-guided replay mechanisms to characterize the stability--plasticity trade-off under synthesis-domain drift. To make ReDIL-GNN predictive rather than purely retrospective, we introduce the \emph{Resynthesis Adaptability Index} (RAI), a pre-adaptation diagnostic that combines adaptation need, source-equivalence recoverability, structural coverage, and update compatibility to estimate whether a new synthesis style is a promising, unnecessary, or risky adaptation target before full incremental training.

In summary, the main contributions of this work are:
\begin{enumerate}
    \item We formulate \emph{resynthesis-domain incremental learning} for circuit GNNs, where semantics-preserving synthesis transformations induce sequential graph-domain shifts while task labels or source identities remain fixed.

    \item We construct a unified evaluation protocol covering classification-style state-of-the-art (SOTA) hardware-security models and representation-learning circuit encoders, with task-native metrics for classifiers and source-equivalence retrieval metrics for embedding models.

    \item We instantiate this predictive adaptation framework with nine incremental update policies: NaiveFT, LwF, Online EWC, MAS, ER, A-GEM, DER++, ER+LwF, and EqReplay, where EqReplay exploits paired source-equivalent implementations across synthesis domains.

    \item We introduce RAI as a lightweight pre-adaptation predictor that connects incoming resynthesis data, frozen-model behavior, structural coverage, and update compatibility to guide when adaptation should be invoked or avoided.
\end{enumerate}

Source code of this work is available at~\cite{anonymous2026domain}.

\begin{table*}[!t]
\centering
\caption{Comparison of ReDIL-GNN with representative related work.
\cmark{} indicates that the capability is explicitly supported, while \xmark{} indicates that it is absent or not the primary formulation.}
\label{tab:related_comparison}
\setlength{\tabcolsep}{2.0pt}
\renewcommand{\arraystretch}{1.10}
\footnotesize
\begin{tabular}{lccccccccc}
\toprule
\textbf{Work / Direction} &
\textbf{Circuit} &
\textbf{Resynth.} &
\textbf{CL} &
\textbf{Forgetting} &
\textbf{Eq. Pairs} &
\textbf{Sup.+Repr.} &
\textbf{Multi-Model} &
\textbf{Pre-Adapt.} &
\textbf{RAI-Guided} \\
&
\textbf{Netlists} &
\textbf{Domains} &
\textbf{Setting} &
\textbf{Measured} &
\textbf{Used} &
\textbf{Metrics} &
\textbf{Audit} &
\textbf{RAI} &
\textbf{Adaptation} \\
\midrule
Circuit GNNs~\cite{zhang2019circuit,alrahis2021gnn,lashen13,yang2022versatile}
& \cmark & \xmark & None & \xmark & \xmark & \xmark & \xmark & \xmark & \xmark \\

Netlist rewriting robustness~\cite{zhao2022gnnrw}
& \cmark & \cmark & None & \xmark & \xmark & \xmark & \xmark & \xmark & \xmark \\

ConVERTS-style representation learning~\cite{chowdhury2023converts}
& \cmark & \cmark & None & \xmark & \cmark & \xmark & \xmark & \xmark & \xmark \\

Adversarial netlist rewriting~\cite{wang2025netdetox}
& \cmark & \cmark & None & \xmark & \cmark & \xmark & \cmark & \xmark & \xmark \\
\midrule
Graph Continual Learning (CL)~\cite{zhou2021overcoming,liu2021overcoming}
& \xmark & \xmark & Task-IL & \cmark & \xmark & \xmark & \xmark & \xmark & \xmark \\

Graph CL adaptation~\cite{guo2025graphkeeper,qiao2025gcal}
& \xmark & \xmark & \emph{Domain-IL} & \cmark & \xmark & \xmark & \xmark & \xmark & \xmark \\
\midrule
Circuit CL benchmark~\cite{karn2026benchmarking}
& \cmark & \xmark & Task-IL & \cmark & \xmark & \xmark & \xmark & \xmark & \xmark \\

Dynamic GNNs for CL~\cite{karn2026dynamic}
& \cmark & \xmark & Task-IL & \cmark & \xmark & \xmark & \cmark & \xmark & \xmark \\

Circuit class incremental learning~\cite{karn2026cilcircuits}
& \cmark & \xmark & Class-IL & \cmark & \xmark & \xmark & \cmark & \xmark & \xmark \\
\midrule
\textbf{ReDIL-GNN (ours)}
& \cmark & \cmark & \textbf{Domain-IL} & \cmark & \cmark & \cmark & \cmark & \cmark & \cmark \\
\bottomrule
\end{tabular}
\end{table*}

\begin{figure}[t]
    \centering
		\includegraphics[scale=0.48, trim = {0.2cm 0cm 0cm 0.5cm}, clip]{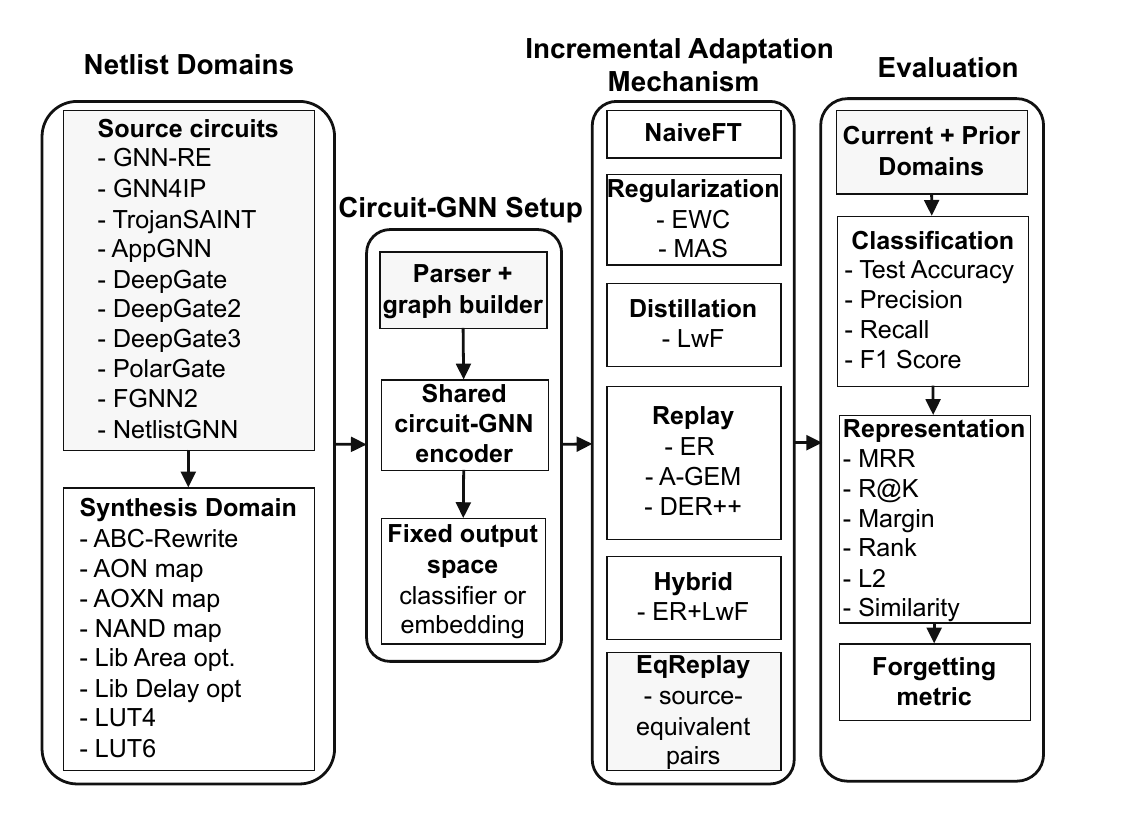}
   \caption{High-level ReDIL-GNN methodology. Source circuits are converted into multiple resynthesis-domain views and parsed into circuit graphs. A shared circuit-GNN encoder with a fixed classifier or embedding space is then adapted through different incremental learning mechanisms. Evaluation is performed on both current and prior domains, using classification metrics for task-specific models, retrieval metrics for representation-learning models, and forgetting metrics to quantify retention.}
    \label{fig:redil_workflow}
\end{figure}

\section{Related Work and Motivation for Ours}
\label{sec:prior-arts}

Prior circuit GNNs address circuit design~\cite{zhang2019circuit}, reverse engineering~\cite{alrahis2021gnn}, Trojan detection~\cite{lashen13}, physical-design prediction~\cite{yang2022versatile}, and functionality-aware representation learning, but they are usually evaluated on fixed graph distributions rather than sequential synthesis-domain shifts.
Resynthesis-aware robustness studies show that equivalent rewrites can degrade predictions~\cite{zhao2022gnnrw}, contrastive methods can use equivalent views for invariance~\cite{chowdhury2023converts}, and adversarial rewriting can exploit local equivalence-preserving transformations against hardware-security GNNs~\cite{wang2025netdetox}.
Continual-learning (CL) methods mitigate forgetting through distillation~\cite{li2016learning}, regularization~\cite{kirkpatrick2017overcoming,aljundi2018memory}, replay~\cite{chaudhry2019tiny}, gradient projection~\cite{chaudhry2019efficient}, and dark replay~\cite{buzzega2020dark}, while graph CL studies forgetting under graph streams~\cite{zhou2021overcoming,liu2021overcoming} or graph-domain shifts~\cite{guo2025graphkeeper,qiao2025gcal}.

Circuit-specific CL studies have established baseline evaluations for gate-level netlists under a task-incremental regime (Task-IL)~\cite{karn2026benchmarking}, showing that static GNNs suffer catastrophic forgetting when learning sequential binary gate and link tasks.
Subsequent work addressed this through dynamic architectural expansion (widening and stacking GNN layers) paired with hyperparameter optimization~\cite{karn2026dynamic}.
Recently, class-incremental learning (CIL) was formulated for circuits to evaluate expanding gate-type label spaces without task identifiers during inference~\cite{karn2026cilcircuits}.
However, all prior circuit CL formulations rely on synthetic task or class splits.

In contrast, ReDIL-GNN focuses on \emph{Domain-IL}, treating resynthesis styles as the incremental domain stream where semantic task labels and architectural heads remain fixed, but the graph topology undergoes semantics-preserving distribution shift.
Unlike static resynthesis-aware training, ReDIL-GNN explicitly measures retention and forgetting; unlike generic graph CL~\cite{zhou2021overcoming,liu2021overcoming,guo2025graphkeeper,qiao2025gcal}, it exploits circuit-specific equivalence between transformed implementations of the same source design through EqReplay.
Furthermore, ReDIL-GNN moves beyond retrospective benchmarking by introducing RAI to actively gate whether, when, and how adaptation should occur before executing costly training updates.
As summarized in Table~\ref{tab:related_comparison}, the resulting framework combines resynthesis-domain adaptation, equivalence-guided replay, retention analysis across diverse SOTA pipelines, and pre-adaptation RAI guidance in a single unified setting.

\begin{table}[t]
\centering
\caption{Resynthesis domains used in ReDIL-GNN.}
\label{tab:resynthesis_domains}
\scriptsize
\setlength{\tabcolsep}{2pt}
\renewcommand{\arraystretch}{1.15}
\begin{tabular}{p{1.6cm}p{6.8cm}}
\toprule
\textbf{Domain name} & \textbf{Meaning / generation style} \\
\midrule
\texttt{Original} 
& Original normalized source netlist, copied without resynthesis. \\
\texttt{ABC-rewrite} 
& Generic ABC rewriting outcome. \\
\texttt{AON} 
& AND/OR/NOT gate-set resynthesis. \\
\texttt{AOXN} 
& AND/OR/XOR/NOT gate-set resynthesis. \\
\texttt{NAND} 
& NAND/NOT gate-set resynthesis. \\
\texttt{Lib-A} 
& Technology-library mapping optimized for area. \\
\texttt{Lib-D} 
& Technology-library mapping optimized for delay. \\
\texttt{LUT4} 
& 4-input LUT mapping. \\
\texttt{LUT6} 
& 6-input LUT mapping. \\
\bottomrule
\end{tabular}
\end{table}

\begin{table*}[t]
\centering
\caption{SOTA GNNs selected in this work. In the \textbf{Pipeline(s)} column, \textcolor{red}{red} denotes classification-based or pairwise-detection models, while \textcolor{blue}{blue} denotes representation/retrieval-based models. Here, \(\rho_m\) denotes the graph-level readout for pipeline \(m\), using the native readout when available or a fixed parameter-free aggregation when no graph-level output is exposed; \(\cos\) denotes cosine similarity between the resulting graph embeddings. Thus, \((\rho_m+\cos)\) denotes graph readout followed by cosine-based source-equivalence retrieval.}
\label{tab:audit_inventory}
\scriptsize
\setlength{\tabcolsep}{3.7pt}
\renewcommand{\arraystretch}{1.08}
\resizebox{\textwidth}{!}{%
\begin{tabular}{lllll}
\toprule
\textbf{Pipeline(s)}
& \textbf{Native objective}
& \textbf{Native graph view}
& \textbf{Retrieval interface}
& \textbf{Datasets} \\
\midrule

\textcolor{red}{GNN-RE~\cite{alrahis2021gnn}}
& Functional reverse engineering
& Gate-level graph
& \(\rho_m+\cos\)
& Arithmetic/interconnected modules \\

\textcolor{red}{GNN4IP~\cite{yasaei2021gnn4ip}}
& Pairwise IP-similarity detection
& Paired gate-level graphs
& Native pairwise head
& Interconnected modules \\

\textcolor{red}{AppGNN~\cite{bucher2022appgnn}}
& Approximation-aware reverse engineering
& Gate-level graph
& \(\rho_m+\cos\)
& Adders and multipliers \\

\textcolor{red}{TrojanSAINT~\cite{lashen2023trojansaint}}
& Trojan-node classification
& Gate-level graph
& \(\rho_m+\cos\)
& TrojanSAINT netlists \\

\textcolor{blue}{DeepGate~\cite{li2022deepgate}, DeepGate2~\cite{shi2023deepgate2},
DeepGate3~\cite{shi2024deepgate3}}
& Gate-function representation
& AIG-style graph
& \(\rho_m+\cos\)
& ITC'99, IWLS'05, EPFL, OpenCore \\

\textcolor{blue}{FGNN2~\cite{wang2024fgnn2}}
& Functional contrastive pretraining
& AIG-style graph
& Embedding cosine
& Synthetic pretraining circuits \\

\textcolor{blue}{PolarGate~\cite{liu2024polargate}}
& Polarity-aware functional representation
& Signed AIG-style graph
& \(\rho_m+\cos\)
& AIGDataset \\

\textcolor{blue}{NetlistGNN/Circuit GNN~\cite{yang2022versatile}}
& Physical-design prediction
& Cell-level circuit graph
& \(\rho_m+\cos\)
& superblue19 \\
\bottomrule
\end{tabular}%
}
\end{table*}

\section{Methodology}
\label{sec:methodology}

\subsection{Overview of ReDIL-GNN}
Figure~\ref{fig:redil_workflow} summarizes the ReDIL-GNN methodology. We consider a stream of resynthesis domains, shown in Table~\ref{tab:resynthesis_domains}, where each incoming domain contains netlists obtained from a new synthesis style while the semantic task remains unchanged. The model is initialized from a multi-domain base checkpoint trained on the seen domains \texttt{Original}, \texttt{ABC-rewrite}, \texttt{AON}, \texttt{NAND}, and \texttt{Lib-D}; \texttt{LUT6} is used only for validation and model selection. The held-out incremental 
stream then introduces \texttt{AOXN}, \texttt{Lib-A}, and \texttt{LUT4} sequentially. Before any adaptation, we also record zero-shot performance on the held-out domains, so that later gains can be separated from robustness already learned during base training. This protocol reflects a realistic deployment setting in which a circuit-GNN may already be trained on several known synthesis flows, but must later adapt to newly encountered tool settings or mapping styles. The goal is therefore not to measure adaptation from a single native distribution, but to evaluate whether a multi-domain circuit-GNN can learn new resynthesis styles while retaining all previously learned styles. At every stage, the same prediction head or embedding space is retained; the model is not given the resynthesis-domain identifier during inference. This makes the setting a domain-incremental learning problem rather than a class-incremental learning problem~\cite{ma2026few,tan2022graph}.

A set of SOTA GNNs used in our work is summarized in Table~\ref{tab:audit_inventory}. Let \(G_i^{(t)}\) denote the implementation of source circuit \(i\) under resynthesis domain \(t\). ReDIL-GNN assumes that the graph distribution changes with \(t\), but the label space or source identity remains fixed. Thus, the goal is not to add new output classes, but to adapt the circuit-GNN pipeline to a newly observed synthesis style while preserving performance on earlier synthesis domains.

\subsection{Model Interfaces: Classification \& Representation Learning}
ReDIL-GNN supports two model families. The first family contains classification-based models, such as node-level reverse-engineering or Trojan-detection GNNs. For these models, the circuit GNN predicts a task label using a shared classifier:
\begin{equation}
    \hat{y}
    =
    f_{\theta}
    \left(
        G_i^{(t)}
    \right),
    \qquad
    \mathcal{L}_{\mathrm{cls}}
    =
    \mathrm{CE}
    \left(
        f_{\theta}
        \left(
            G_i^{(t)}
        \right),
        y_i
    \right).
    \label{eq:redil_classification_loss}
\end{equation}
For example, in a GNN-RE-style functional classification task~\cite{alrahis2021gnn}, the output classes remain the same across resynthesis styles, but the underlying gate-level graph may change substantially. The adaptation objective is therefore to learn the current graph distribution while preserving the previously learned decision boundary.

The second family contains representation-learning models, such as the DeepGate family~\cite{li2022deepgate,shi2023deepgate2,shi2024deepgate3}, FGNN2, PolarGate, and NetlistGNN. For a pipeline \(m\), let \(\rho_m(\cdot)\) denote its graph-level readout: the native graph readout when the model exposes one, or a fixed parameter-free aggregation of node embeddings when no graph-level output is available. Given node or graph representations produced by encoder \(e_{\theta}\), we write the graph embedding as
\begin{equation}
    \mathbf{z}_{i,m}^{(t)}
    =
    \rho_m
    \left(
        e_{\theta}
        \left(
            G_i^{(t)}
        \right)
    \right).
    \label{eq:redil_readout}
\end{equation}
The notation \(\rho_m+\cos\), used in Table~\ref{tab:audit_inventory}, therefore means applying the readout \(\rho_m\) and then comparing the resulting graph embeddings with cosine similarity:
\begin{equation}
    s_m
    \left(
        G_i^{(t)},
        G_j^{(0)}
    \right)
    =
    \cos
    \left(
        \mathbf{z}_{i,m}^{(t)},
        \mathbf{z}_{j,m}^{(0)}
    \right).
    \label{eq:redil_readout_cosine}
\end{equation}
A transformed query \(G_i^{(t)}\) is compared against an \texttt{Original}-domain gallery, and the correct source \(G_i^{(0)}\) should rank ahead of unrelated circuits. Thus, for embedding models, we report source-equivalence retrieval metrics such as Recall@K, MRR, median rank, positive-pair similarity, \(L_2\) displacement, and source margin, rather than treating binary F1 as the primary metric.

\subsection{Incremental Adaptation Mechanisms}
At each stage, ReDIL-GNN updates the model using one of nine adaptation mechanisms (see Figure~\ref{fig:redil_workflow}). The simplest baseline is naive fine-tuning, which optimizes only the current-domain loss and provides a lower-bound measure of forgetting. Distillation-based learning without forgetting (LwF)~\cite{li2016learning} keeps a frozen copy of the previous model and penalizes deviations from its predictions or similarity scores. Parameter-regularization methods, Online EWC~\cite{kirkpatrick2017overcoming,schwarz2018progress} and MAS~\cite{aljundi2018memory}, estimate which parameters are important for previous domains and discourage large changes to them.

Replay-based methods retain examples from earlier domains. ER~\cite{rolnick2019experience,chaudhry2019tiny} maintains a bounded memory of previous-domain examples and mixes them with current-domain data. DER++~\cite{buzzega2020dark} additionally stores historical logits or scores, allowing the model to preserve both labels and previous responses. A-GEM~\cite{chaudhry2019efficient} uses replay examples to project gradients when the current update would increase old-domain loss. ER+LwF combines replay with distillation~\cite{rolnick2019experience,li2016learning}.

EqReplay is the circuit-specific mechanism in ReDIL-GNN. It uses source-equivalent pairs across adjacent resynthesis domains, treating \(G_i^{(t)}\) and \(G_i^{(t-1)}\) as two implementations of the same circuit rather than as unrelated replay samples. For example, after adapting from \texttt{Lib-A} to \texttt{LUT4}, EqReplay can pair the \texttt{LUT4} implementation of source circuit \(i\) with its previously learned \texttt{Lib-A} implementation, then penalize inconsistent class predictions for classification models or misaligned embeddings and similarity scores for representation-learning models.

\subsection{Stage-Wise Training and Evaluation}
The same stage-wise protocol is used for all models. First, a shared base checkpoint is trained on the base domains. Then, for each incoming resynthesis domain, the chosen mechanism updates the model using the current domain and any allowed memory, teacher, importance estimate, or source-equivalent replay examples. After the update, the model is evaluated on the current domain and all previously observed domains. This produces a stage-by-domain performance matrix, from which we compute final performance, worst-domain performance, and forgetting.

The evaluation separates adaptation from retention. Current-domain performance measures plasticity, i.e., how well the model learns the new resynthesis style. Prior-domain performance measures stability, i.e., how much old-domain performance is preserved. Forgetting for a domain is computed as the drop between the best score previously achieved on that domain and the score after the final incremental stage:
\begin{equation}
    F_d
    =
    \max_{\tau \leq T}
    M_{\tau,d}
    -
    M_{T,d},
    \label{eq:redil_forgetting}
\end{equation}
where \(M_{\tau,d}\) is the metric value on domain \(d\) after stage \(\tau\), and \(T\) is the final stage. For classification-based models, \(M_{\tau,d}\) can be macro-F1, balanced accuracy, or task-specific detection performance. For representation-learning models, \(M_{\tau,d}\) is primarily MRR, Recall@K, or source margin. This allows ReDIL-GNN to compare classification-based SOTA models and embedding-based SOTA models without forcing all models into a single metric type.

\subsection{Resynthesis Adaptability Index}
\label{sec:rai}

RAI uses the frozen base model, a small probe set from the incoming resynthesis domain, and a small retention probe from the base domains. Its purpose is to estimate whether adapting to the new synthesis style is likely to be useful, unnecessary, or risky.

Let \(m\) denote the circuit-GNN model, \(c\) denote the incremental-learning mechanism, and \(d_t\) denote the incoming resynthesis domain at stage \(t\). Let \(\mathcal{Q}_t\) be a small probe set from \(d_t\), and let \(\mathcal{B}\) be a small probe set from the base or retention domains. We define RAI using four normalized components:
\begin{equation}
    \mathrm{RAI}_{m,c,t}
    =
    \left(
        N_{m,t}
        \cdot
        R_{m,t}
        \cdot
        S_t
        \cdot
        C_{m,c,t}
    \right)^{1/4}.
    \label{eq:rai}
\end{equation}
Each term lies in \([0,1]\), and larger values indicate that the incoming domain is a better candidate for safe and useful adaptation. The geometric form makes the score conservative: RAI is high only when adaptation is needed, the new domain is recoverable, the training data are structurally covered, and the update is unlikely to damage retention.

The first term, \(N_{m,t}\), measures \emph{adaptation need}. It is high when the frozen base model performs poorly on the incoming-domain probe:
\begin{equation}
    N_{m,t}
    =
    1
    -
    \widehat{M}_{m,0}
    \left(
        \mathcal{Q}_t
    \right),
    \label{eq:rai_need}
\end{equation}
where \(\widehat{M}_{m,0}\) is the normalized zero-shot score of the base model. For classification models, \(M\) can be macro-F1, balanced accuracy, or task-specific F1; for embedding models, \(M\) can be Recall@1 or MRR. Thus, a low \(N_{m,t}\) indicates that the base model already generalizes to the new resynthesis style, while a high \(N_{m,t}\) indicates that adaptation may be needed.

The second term, \(R_{m,t}\), measures \emph{source-equivalence recoverability}. It asks whether the base model still recognizes that two different implementations of the same source circuit are related. For embedding-based models, this is measured using the similarity gap between the correct source-equivalent pair and the closest incorrect source:
\begin{equation}
    R_{m,t}
    =
    \mathrm{norm}
    \left(
        s(G_i^{(t)},G_i^{(0)})
        -
        \max_{j \neq i}
        s(G_i^{(t)},G_j^{(0)})
    \right).
    \label{eq:rai_recoverability_embedding}
\end{equation}
For classification models, \(R_{m,t}\) is measured as the prediction consistency between \(G_i^{(t)}\) and \(G_i^{(0)}\). A high recoverability score means that the model has not completely lost the functional connection between the transformed graph and its original implementation.

The third term, \(S_t\), measures \emph{structural coverage}. It compares graph descriptors from the incoming resynthesis domain with the descriptors of the base domains:
\begin{equation}
    S_t
    =
    \mathrm{sim}
    \left(
        \psi(\mathcal{Q}_t),
        \psi(\mathcal{B})
    \right),
    \label{eq:rai_structural_coverage}
\end{equation}
where \(\psi(\cdot)\) denotes lightweight graph statistics such as node count, edge count, gate or cell histogram, fan-in/fan-out statistics, depth, PI/PO counts, LUT/cell-type ratios, or frozen-model embedding summaries. A high \(S_t\) indicates that the incoming resynthesis style is structurally close to at least one previously seen synthesis style, while a low \(S_t\) suggests that the new domain lies outside the training distribution.

The fourth term, \(C_{m,c,t}\), measures \emph{update compatibility}. It estimates whether a small probe update on the incoming domain agrees with the base-domain objective:
\begin{equation}
    C_{m,c,t}
    =
    \mathrm{norm}
    \left(
        \cos
        \left(
            g_{m,c,t}^{\mathrm{new}},
            g_{m}^{\mathrm{base}}
        \right)
    \right),
    \label{eq:rai_compatibility}
\end{equation}
where \(g_{m,c,t}^{\mathrm{new}}\) is the probe gradient induced by the incoming domain under mechanism \(c\), and \(g_{m}^{\mathrm{base}}\) is the gradient on the base-domain retention probe. A high value means that the new-domain update is compatible with the base domains; a low value indicates that adaptation may improve the current domain but cause forgetting.

RAI is used as a decision guide rather than as a post-hoc result metric. If \(N_{m,t}\) is low, adaptation can be skipped because the base model already handles the new resynthesis style. If \(N_{m,t}\) is high and RAI is also high, incremental learning is expected to be beneficial. If \(N_{m,t}\) is high but RAI is low, adaptation is risky: the model either lacks source-equivalence recoverability, the new graphs are structurally far from the base domains, or the update direction conflicts with retention. In such cases, retention-heavy mechanisms such as replay, DER++, ER+LwF, A-GEM, or EqReplay should be preferred over naive fine-tuning.

In summary, RAI converts ReDIL-GNN from only an evaluation benchmark into a predictive framework. It connects the incoming resynthesis data, the frozen model behavior, and the expected stability of the update before full incremental training is performed. This allows a designer to decide whether a new synthesis style should be learned, whether the existing base model is already sufficient, or whether adaptation may cause unacceptable forgetting. Because RAI is the geometric mean of four normalized terms, its value also lies in \([0,1]\): \(\mathrm{RAI}=0\) is the theoretical minimum and occurs when at least one required condition is absent, while \(\mathrm{RAI}=1\) is the ideal maximum and indicates that adaptation is needed, recoverable, structurally covered, and compatible with retention. Thus, values close to zero suggest that incremental learning should be skipped or applied only with strong retention safeguards, whereas values close to one suggest that the incoming resynthesis domain is a strong candidate for safe and beneficial adaptation.

\begin{table}[t]
\centering
\caption{Representative resynthesis statistics for GNN-RE. 
Rows report the eight resynthesized domains; \texttt{ABC-rewrite} denotes the collapsed ABC rewriting outcome. 
\emph{Comp.} denotes weakly connected components and \(G_{L_1}\) denotes gate-composition drift from the original netlist. 
The \(\Delta\) columns report percentage change with respect to the original netlist.}
\label{tab:gnnre_resynthesis_statistics_compact}
\scriptsize
\setlength{\tabcolsep}{2.2pt}
\renewcommand{\arraystretch}{1.04}
\resizebox{\columnwidth}{!}{%
\begin{tabular}{lrrrrrrrr}
\toprule
\textbf{Outcome}
& \textbf{Nodes}
& \textbf{Edges}
& \textbf{Depth}
& \textbf{Comp.}
& \(\boldsymbol{G_{L_1}}\)
& \(\boldsymbol{\Delta N}\)
& \(\boldsymbol{\Delta E}\)
& \(\boldsymbol{\Delta D}\) \\
\midrule
\texttt{Original}     & 2830.5 & 9416.4 & 20.3 & 1.2    & 0.000 & 0.0  & 0.0   & 0.0 \\
\texttt{ABC-rewrite} & 2227.9 & 2122.3 & 23.7 & 1002.5 & 1.990 & 35.3 & -51.3 & 31.0 \\
\texttt{AON}         & 2436.9 & 2430.6 & 36.4 & 1002.5 & 1.990 & 41.2 & -47.1 & 64.7 \\
\texttt{AOXN}        & 2003.5 & 1589.5 & 35.2 & 1002.5 & 1.990 & 26.7 & -56.9 & 57.1 \\
\texttt{NAND}        & 2380.7 & 2261.2 & 44.5 & 1002.5 & 1.990 & 52.1 & -40.1 & 109.1 \\
\texttt{Lib-A}       & 1760.0 & 5444.9 & 22.0 & 1000.2 & 1.388 & 19.4 & -19.6 & 8.7 \\
\texttt{Lib-D}       & 1794.3 & 5581.5 & 22.6 & 1000.3 & 1.397 & 22.0 & -18.8 & 7.5 \\
\texttt{LUT4}        & 1373.0 & 778.0  & 12.0 & 1002.6 & 1.990 & -10.2 & -83.2 & -45.6 \\
\texttt{LUT6}        & 2026.4 & 1057.1 & 7.6  & 1599.5 & 1.990 & 8.1 & -80.6 & -59.1 \\
\midrule
\textbf{Mean (resyn.)}
& \textbf{2000.3}
& \textbf{2658.1}
& \textbf{25.5}
& \textbf{1076.6}
& \textbf{1.841}
& \textbf{24.3}
& \textbf{-49.7}
& \textbf{21.7} \\
\bottomrule
\end{tabular}%
}
\end{table}

\begin{table}[t]
\centering
\caption{Summary of resynthesis statistics across reproduced SOTA works. 
Each row is the mean over the eight resynthesis outcomes. 
GNN4IP is omitted because it reuses the GNN-RE resynthesized graph artifacts; DeepGate, DeepGate2, and DeepGate3 are summarized together because they use the common DeepGate-family / DeepGate2-dataset structural statistics. 
The complete per-domain statistics are provided in the supplementary document released with the code \cite{anonymous2026domain}.}
\label{tab:sota_resynthesis_statistics_summary}
\scriptsize
\setlength{\tabcolsep}{2.0pt}
\renewcommand{\arraystretch}{1.04}
\resizebox{\columnwidth}{!}{%
\begin{tabular}{lrrrrrrrr}
\toprule
\textbf{Work}
& \textbf{Nodes}
& \textbf{Edges}
& \textbf{Depth}
& \textbf{Comp.}
& \(\boldsymbol{G_{L_1}}\)
& \(\boldsymbol{\Delta N}\)
& \(\boldsymbol{\Delta E}\)
& \(\boldsymbol{\Delta D}\) \\
\midrule
AppGNN          & 1621.8 & 4005.5 & 10.3 & 1209.8 & 1.667 & -4.9  & -32.6 & -20.9 \\
DeepGate-family & 209.7  & 278.3  & 10.5 & 12.4   & 0.505 & -57.5 & -35.4 & 197.5 \\
FGNN2           & 16.4   & 23.8   & 4.0  & 2.0    & 0.421 & -23.3 & -29.8 & -34.6 \\
NetlistGNN      & 20.8   & 22.9   & 5.2  & 5.1    & 1.307 & -43.6 & -51.0 & -51.0 \\
PolarGate       & 159.9  & 207.0  & 11.2 & 15.7   & 0.243 & -15.6 & -17.2 & -27.6 \\
GNN-RE          & 2000.3 & 2658.1 & 25.5 & 1076.6 & 1.841 & 24.3  & -49.7 & 21.7 \\
TrojanSAINT     & 6943.8 & 3751.5 & 7.4  & 3325.3 & 1.963 & 46.0  & -65.7 & -59.0 \\
\bottomrule
\end{tabular}%
}
\end{table}

\section{Results}
\label{sec:results}

\subsection{Resynthesis Statistics}
\label{subsec:resynthesis_statistics}

To document the structural effect of the resynthesis front end, we report compact graph-level statistics for the generated source-equivalent domains. The full supplementary table records nodes, edges, depth, fan-in/fan-out, density, connected components, node/gate-type entropy, gate-composition drift, raw graph-statistic drift, and percentage changes relative to the original netlist; it also states that GNN4IP reuses the GNN-RE resynthesized graph artifacts and that DeepGate, DeepGate2, and DeepGate3 share the same DeepGate-family structural statistics.  Table~\ref{tab:gnnre_resynthesis_statistics_compact} shows a representative single-column view for GNN-RE over the eight resynthesis outcomes, while Table~\ref{tab:sota_resynthesis_statistics_summary} reports one summary row per reproduced SOTA work; the complete per-transformation statistics are provided in the supplementary document released with the code \cite{anonymous2026domain}. These statistics confirm that the resynthesis flow is not a superficial renaming of the input netlists: for example, GNN-RE shows large edge reductions for LUT mappings and substantial component/gate-composition changes, while the summary table shows that the same phenomenon appears across classification and embedding-oriented circuit-GNN pipelines.

\begin{figure}[!t]
    \centering
		\includegraphics[scale=0.5, trim = {0cm 0cm 0cm 0cm}, clip]{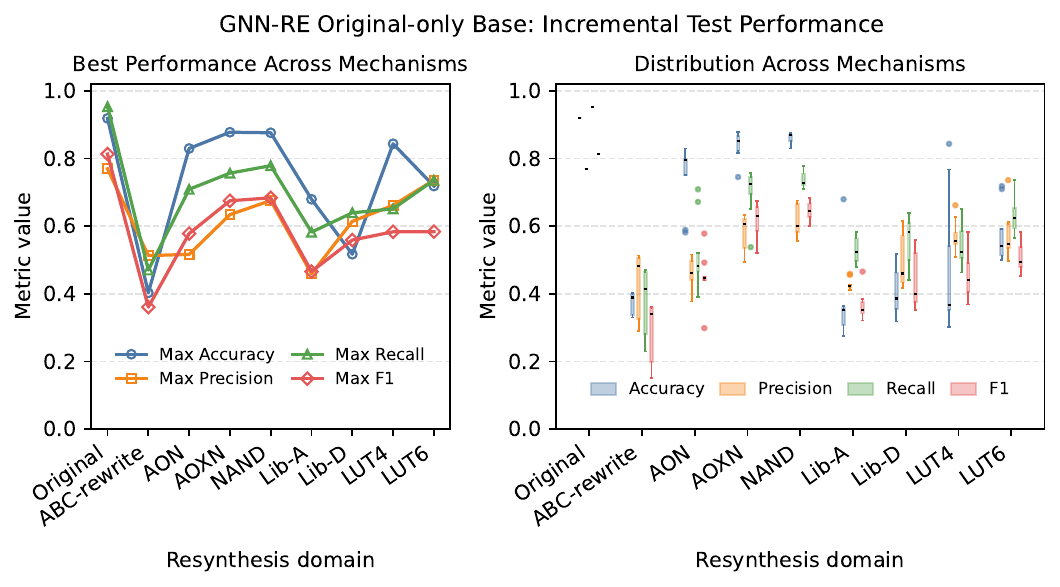}
   \caption{\emph{Domain-IL} for GNN-RE where base model is trained only with original netlist. }
    \label{fig:original_only_base_gnnre}
\end{figure}

\begin{figure}[!t]
    \centering
		\includegraphics[scale=0.45, trim = {0cm 0cm 0cm 0cm}, clip]{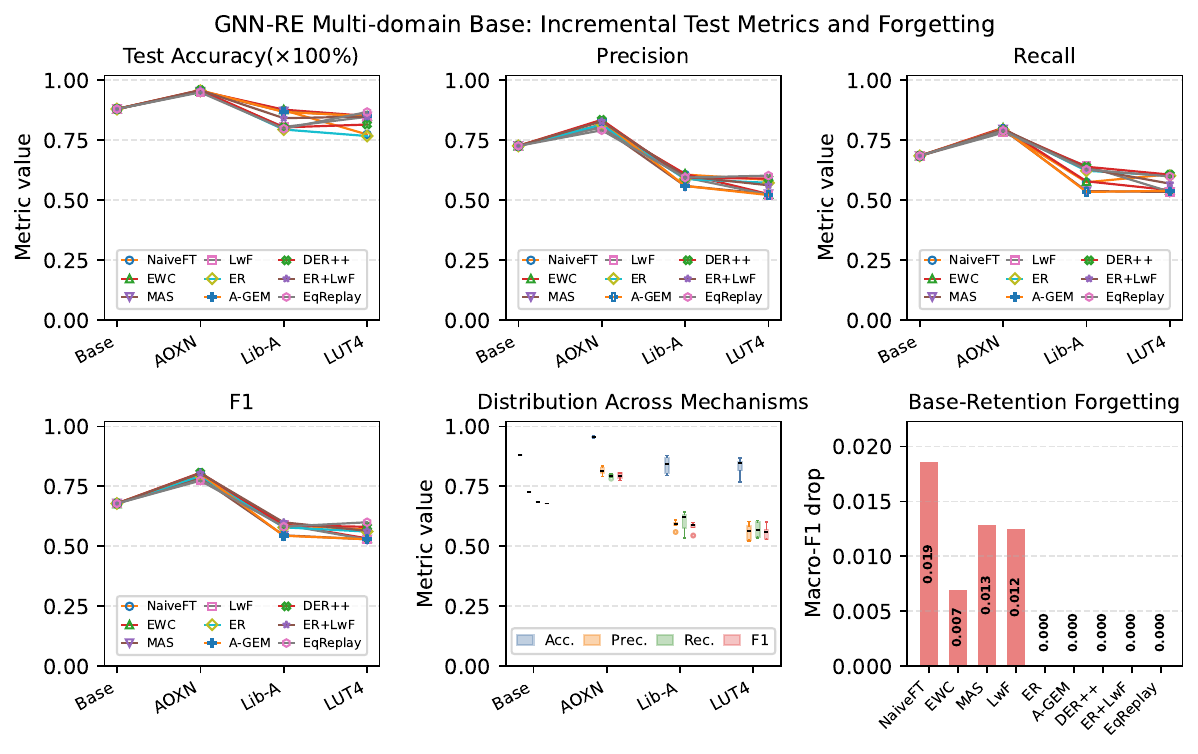}
   \caption{\emph{Domain-IL} for GNN-RE where base model is trained with Original, Rewrite, AON, NAND, Lib-D and incrementally learned AOXN, Lib-A, and LUT4 with LUT6 set aside as validation during training. }
    \label{fig:gnnre_incremental_stage}
\end{figure}

\begin{figure}[!t]
    \centering
		\includegraphics[scale=0.55, trim = {0cm 0cm 0cm 0cm}, clip]{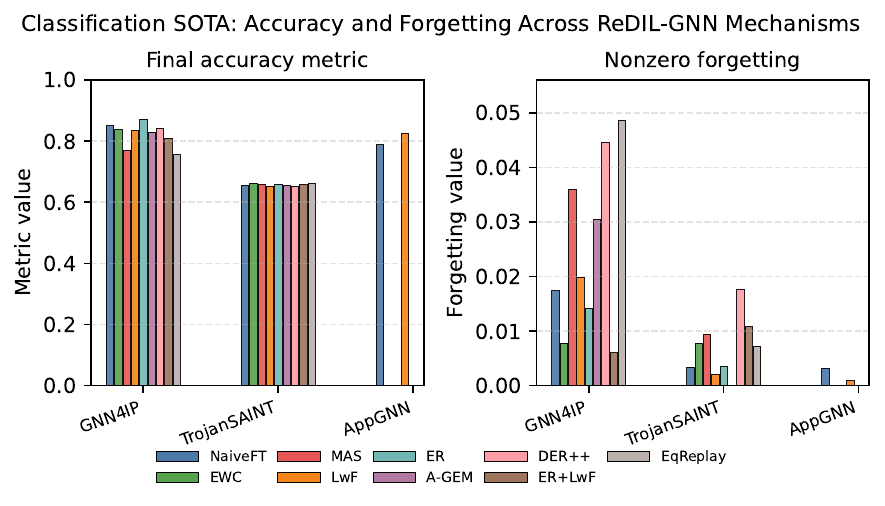}
   \caption{\emph{Domain-IL} performance for GNN4IP, TrojanSAINT, and AppGNN where training setup is same as Fig.~\ref{fig:gnnre_incremental_stage}. }
    \label{fig:classification_sota_performance}
\end{figure}

\subsection{\emph{Domain-IL} Results for Classification-Based SOTA GNNs}
\label{sec:results_classification_sota}

Figure~\ref{fig:original_only_base_gnnre} first shows a strict baseline where GNN-RE is trained only on the \texttt{Original} netlists and then adapted sequentially to each resynthesis style. The left panel reports the maximum accuracy, precision, recall, and F1 obtained across \emph{Domain-IL} mechanisms, while the right panel shows the distribution of these metrics across those mechanisms. This setting exposes the severity of resynthesis-induced domain shift: when the base model has seen only the native netlist distribution, performance varies substantially as the stream progresses through \texttt{ABC-rewrite}, \texttt{AON}, \texttt{AOXN}, \texttt{NAND}, \texttt{Lib-A}, \texttt{Lib-D}, \texttt{LUT4}, and \texttt{LUT6}. The spread in the boxplots further indicates that the choice of adaptation mechanism matters, because different methods respond differently to each synthesis transformation. Thus, this original-only experiment serves as the lower-bound deployment scenario and motivates using a stronger multi-domain base model for the main ReDIL-GNN evaluation. Similar behavior applies to GNN4IP, TrojanSAINT, and AppGNN which are omitted here due to paper space constraints and they are available in our release \cite{anonymous2026domain}.

Figure~\ref{fig:gnnre_incremental_stage} reports the main GNN-RE incremental experiment, where the base model is trained on multiple seen domains and then adapted to the held-out stream \texttt{AOXN}~$\rightarrow$~\texttt{Lib-A}~$\rightarrow$~\texttt{LUT4}. The first four panels track accuracy, macro precision, macro recall, and macro F1 after each learned resynthesis domain, while the fifth panel summarizes the distribution across mechanisms and the sixth panel reports base-retention forgetting. Compared with the original-only baseline, the multi-domain base produces a more stable starting point because the model has already learned several synthesis-induced graph variations before the incremental stream begins. Nevertheless, the trajectories still show that adaptation is nontrivial: current-domain performance and base-domain retention do not improve uniformly across mechanisms. Replay-based and equivalence-guided mechanisms are generally more stable, while methods without explicit retention constraints are more prone to drops in the retention panel. This figure therefore supports the central premise of ReDIL-GNN: resynthesis-domain adaptation must be evaluated jointly in terms of plasticity on the current domain and stability on previously learned domains.

Figure~\ref{fig:classification_sota_performance} consolidates the classification-based SOTA results for GNN4IP, TrojanSAINT, and AppGNN. The first panel compares the final accuracy-like metric for each ReDIL-GNN mechanism, while the second panel compares the corresponding nonzero forgetting values. This consolidated view shows that no single mechanism dominates all classification-style models, but it also highlights a consistent trend: mechanisms that preserve prior knowledge through replay, distillation, or equivalence-aware alignment tend to offer a better stability--plasticity trade-off than naive adaptation alone. Due to space constraints, we include detailed stage-wise plots only for GNN-RE in the paper; the corresponding detailed plots and CSV outputs for the other classification-based GNNs are provided in the release \cite{anonymous2026domain}.

The main {\em takeaway} from the classification results is that resynthesis adaptation is not simply a matter of improving the current-domain score. 
The original-only experiment shows the lower-bound deployment scenario: when the base model has seen only the native graph distribution, each new synthesis style can reshape the decision boundary in a different way. 
The multi-domain-base setting reduces this shock but does not eliminate the stability--plasticity trade-off, because a mechanism that improves the incoming domain can still damage retention on earlier domains. 
Thus, for classification-based circuit GNNs, the practical lesson is to train the base model with diverse synthesis views when possible and to prefer retention-aware updates whenever the incoming style is not already covered.

\begin{figure}[!t]
    \centering
		\includegraphics[scale=0.45, trim = {0cm 0cm 0cm 0cm}, clip]{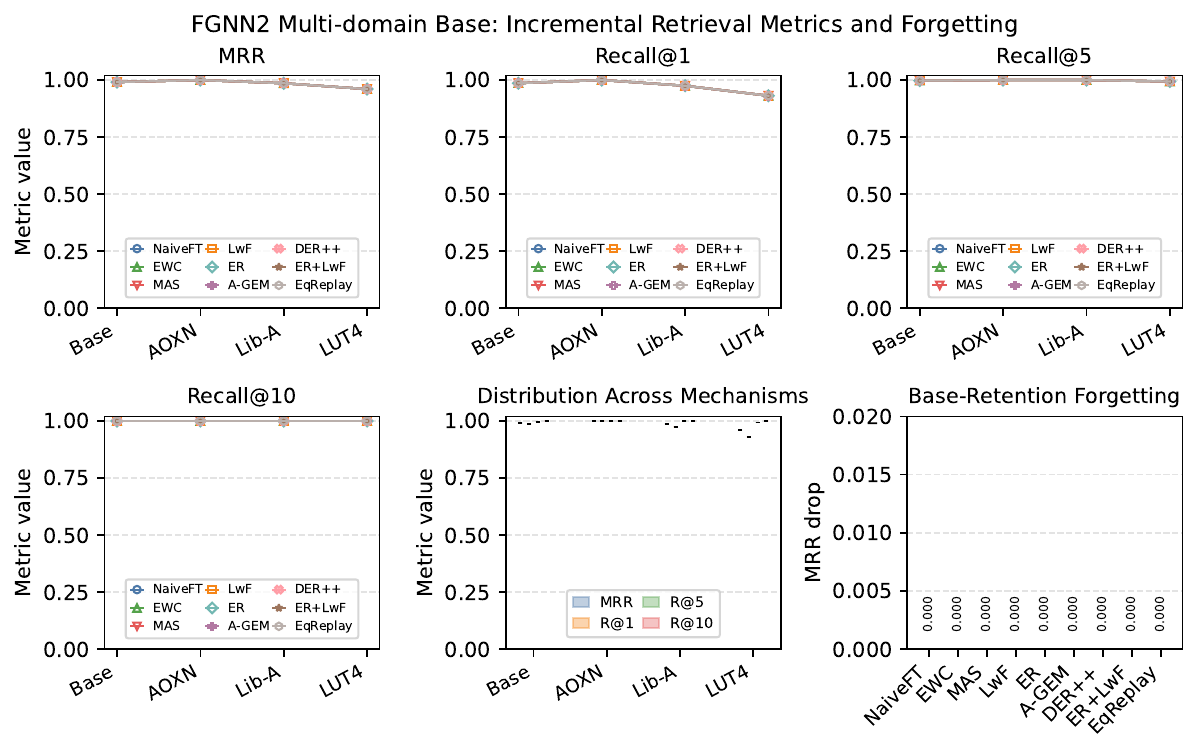}
   \caption{\emph{Domain-IL} for FGNN2 where base model is trained with Original, Rewrite, AON, NAND, Lib-D and incrementally learned AOXN, Lib-A, and LUT4 with LUT6 set aside as validation during training. }
    \label{fig:fgnn2_incremental_stage}
\end{figure}

\begin{figure}[!t]
    \centering
		\includegraphics[scale=0.55, trim = {0cm 0cm 0cm 0cm}, clip]{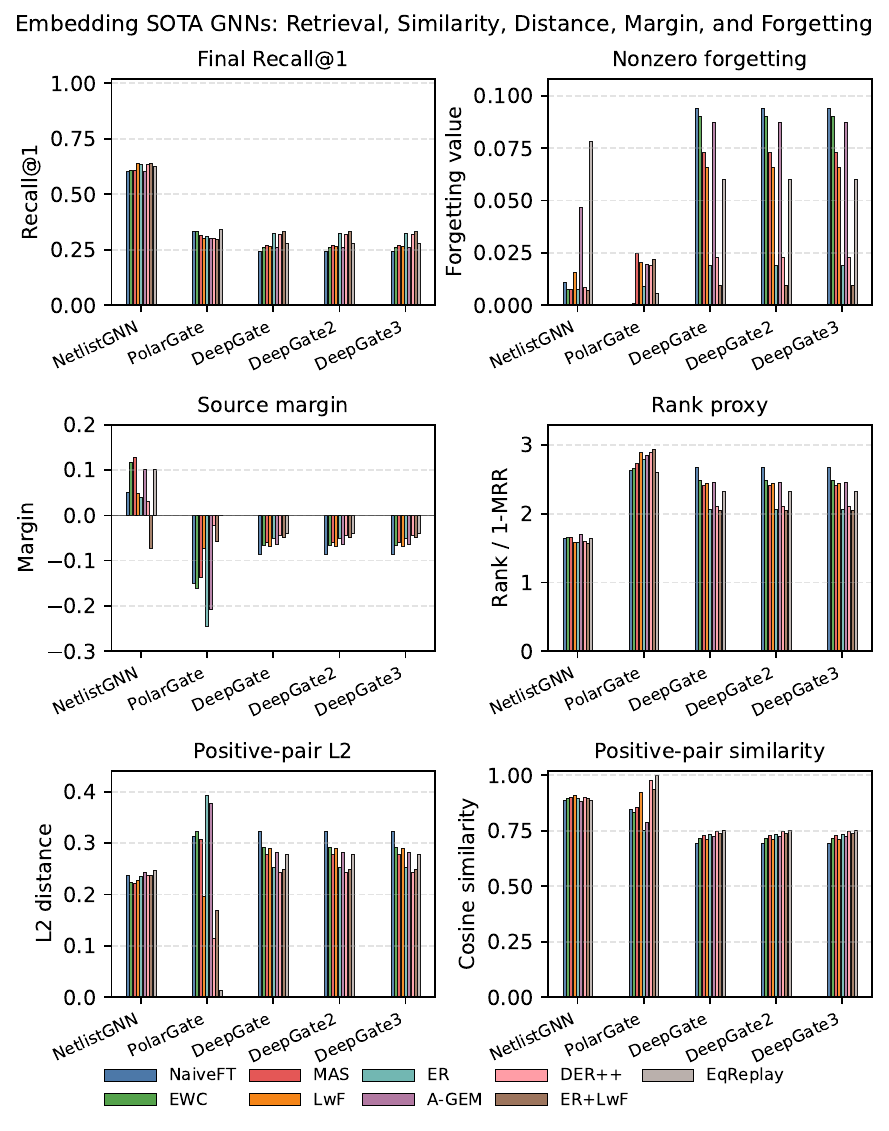}
   \caption{\emph{Domain-IL} performance for FGNN2, Netlist/Circuit GNN, PolarGate, DeepGate, DeepGate2, and DeepGate3 where training setup is same as Fig.~\ref{fig:fgnn2_incremental_stage}. }
    \label{fig:embedding_sota_six_metric}
\end{figure}

\subsection{\emph{Domain-IL} Results for Embedding-Based SOTA GNNs}
\label{sec:results_embedding_sota}

Figure~\ref{fig:fgnn2_incremental_stage} shows the detailed embedding-based \emph{Domain-IL} behavior using FGNN2 as the representative case. FGNN2 is selected because it provides the strongest retrieval behavior among the embedding-based SOTA GNNs, and the figure tracks MRR, Recall@1, Recall@5, Recall@10, the across-mechanism distribution, and base-retention MRR forgetting as the model moves from the base checkpoint to \texttt{AOXN}, \texttt{Lib-A}, and \texttt{LUT4}. The results show that the FGNN2 representation is highly stable under the evaluated resynthesis stream: retrieval metrics remain close to saturation for most mechanisms, and the forgetting panel indicates limited loss of base-domain retrieval ability after learning the final \texttt{LUT4} domain. This suggests that functional contrastive representations and source-equivalence retrieval can be more naturally aligned with resynthesis robustness than strict task-label prediction, because the model is not forced to preserve an exact gate-level decision boundary across structurally different but functionally equivalent graphs. Still, the distribution panel shows that not all mechanisms are identical; replay and hybrid replay-distillation mechanisms provide stable retention, while methods without explicit memory or alignment can show larger variation across stages. Due to space constraints, we report the full stage-wise trajectory only for FGNN2, while the corresponding detailed plots and CSV files for the remaining embedding-based models are included in the release \cite{anonymous2026domain}.

Figure~\ref{fig:embedding_sota_six_metric} consolidates the final embedding-based results after the last incremental domain, \texttt{LUT4}, is learned. The figure compares remaining embedding-based SOTA GNNs including Netlist/Circuit GNN, PolarGate, DeepGate, DeepGate2, and DeepGate3 using Recall@1, nonzero forgetting, source margin, rank proxy, positive-pair $L_2$ distance, and positive-pair similarity. The consolidated view shows that embedding-based SOTA GNNs are not equally robust to resynthesis-domain adaptation: FGNN2 remain the strongest retrieval comapring Figure~\ref{fig:fgnn2_incremental_stage} with \ref{fig:embedding_sota_six_metric}, Netlist/Circuit GNN achieves moderate source-identification performance, and the DeepGate-family models remain more sensitive to the resynthesis stream despite using representation-level retrieval metrics. PolarGate shows high positive-pair similarity for some mechanisms, but its Recall@1 and margin behavior indicate that high cosine similarity alone is insufficient when hard negatives remain close to the true source. The forgetting subplot further shows that retention is mechanism-dependent; replay and hybrid methods generally reduce retrieval degradation, while EqReplay can improve equivalence alignment in some models but does not uniformly dominate every metric. These results motivate the use of multiple representation-centric metrics rather than a single retrieval score: Recall@1 captures source-identification success, margin and rank capture separation from hard negatives, $L_2$ and similarity capture embedding alignment, and forgetting quantifies retention after sequential resynthesis adaptation.

The embedding results suggest a different deployment lesson from the classification results. Representation models can be more naturally aligned with resynthesis robustness because they are asked to preserve source identity rather than reproduce a fixed gate-level decision boundary. 
However, high positive-pair similarity alone is not sufficient: a useful embedding must also separate the correct source from hard negatives, preserve rank, and avoid forgetting previous domains. Therefore, source-equivalence retrieval under resynthesis should be evaluated with a metric bundle, including Recall@K, MRR, margin, distance, similarity, and forgetting, rather than with a single retrieval score.

\begin{table}[t]
\centering
\caption{Compact best--worst pre-adaptation RAI summary for classification-based SOTA GNNs. 
For tied cases, a single row is shown. 
For non-tied cases, the best and worst mechanisms are shown in separate rows.}
\label{tab:classification_sota_rai_best_worst_compact}
\scriptsize
\setlength{\tabcolsep}{2.5pt}
\renewcommand{\arraystretch}{1.04}
\resizebox{\columnwidth}{!}{%
\begin{tabular}{lllccccc}
\toprule
\textbf{Model}
& \textbf{Incoming}
& \textbf{Case / Mechanism}
& \textbf{$N$}
& \textbf{$R$}
& \textbf{$S$}
& \textbf{$C$}
& \textbf{RAI} \\
\midrule

\multirow{3}{*}{GNN-RE} 
& \texttt{AOXN} 
& \textsc{Tie: All} 
& 0.101 & 0.989 & 1.000 & 0.968 & 0.558 \\
\cmidrule{2-7}
& \multirow{2}[1]{*}{\texttt{Lib-A}} 
& Best: \textsc{EqReplay} 
& 0.364 & 0.877 & 0.952 & 0.654 & \textbf{0.668} \\
& & Worst: \textsc{NaiveFT} 
& 0.329 & 0.852 & 0.952 & 0.464 & 0.593 \\
\cmidrule{2-7}
& \multirow{2}[1]{*}{\texttt{LUT4}} 
& Best: \textsc{DER++} 
& 0.367 & 0.904 & 0.740 & 0.563 & \textbf{0.610} \\
& & Worst: \textsc{MAS} 
& 0.263 & 0.902 & 0.740 & 0.447 & 0.530 \\

\midrule

\multirow{3}{*}{GNN4IP} 
& \texttt{AOXN} 
& \textsc{Tie: All} 
& 0.113 & 0.782 & 1.000 & 0.990 & 0.544 \\
\cmidrule{2-7}
& \texttt{Lib-A} 
& \textsc{Tie: All} 
& 0.219 & 0.725 & 0.924 & 0.931 & 0.608 \\
\cmidrule{2-7}
& \multirow{2}[1]{*}{\texttt{LUT4}} 
& Best: \textsc{EWC} 
& 0.460 & 0.404 & 0.716 & 0.863 & \textbf{0.582} \\
& & Worst: \textsc{ER} 
& 0.451 & 0.322 & 0.716 & 0.345 & 0.435 \\

\midrule

\multirow{3}{*}{AppGNN} 
& \texttt{AOXN} 
& \textsc{Tie: All} 
& 0.383 & 0.961 & 0.995 & 0.963 & 0.771 \\
\cmidrule{2-7}
& \texttt{Lib-A} 
& \textsc{Tie: All} 
& 0.522 & 0.916 & 0.955 & 0.891 & 0.799 \\
\cmidrule{2-7}
& \multirow{2}[1]{*}{\texttt{LUT4}} 
& Best: \textsc{EqReplay} 
& 0.714 & 0.879 & 0.769 & 0.859 & \textbf{0.802} \\
& & Worst: \textsc{ER+LwF} 
& 0.718 & 0.879 & 0.769 & 0.815 & 0.793 \\

\midrule

\multirow{3}{*}{TrojanSAINT} 
& \texttt{AOXN} 
& \textsc{Tie: All} 
& 0.067 & 1.000 & 1.000 & 0.618 & 0.452 \\
\cmidrule{2-7}
& \texttt{Lib-A} 
& \textsc{Tie: All} 
& 0.727 & 0.953 & 1.000 & 0.480 & 0.759 \\
\cmidrule{2-7}
& \texttt{LUT4} 
& \textsc{Tie: All} 
& 0.375 & 0.976 & 0.978 & 0.804 & 0.732 \\

\bottomrule
\end{tabular}%
}
\end{table}

\begin{table}[t]
\centering
\caption{Compact best--worst pre-adaptation RAI summary for the GNN-RE original-only-base protocol.
The base model is trained only on \texttt{Original}, and each resynthesis style is then introduced sequentially. 
For tied cases, a single row is shown; for non-tied cases, best and worst mechanisms are shown separately.}
\label{tab:gnnre_original_only_rai_best_worst_compact}
\scriptsize
\setlength{\tabcolsep}{2.5pt}
\renewcommand{\arraystretch}{1.04}
\resizebox{\columnwidth}{!}{%
\begin{tabular}{llccccc}
\toprule
\textbf{Incoming}
& \textbf{Case / Mechanism}
& \textbf{$N$}
& \textbf{$R$}
& \textbf{$S$}
& \textbf{$C$}
& \textbf{RAI} \\
\midrule

\texttt{ABC-rewrite}
& \textsc{Tie: All}
& 0.798 & 0.736 & 0.000 & 0.583 & 0.001 \\

\cmidrule{1-7}
\multirow{2}[1]{*}{\texttt{AON}}
& Best: \textsc{LwF}
& 0.757 & 0.733 & 0.234 & 0.467 & \textbf{0.496} \\
& Worst: \textsc{DER++}
& 0.740 & 0.721 & 0.234 & 0.453 & 0.488 \\

\cmidrule{1-7}
\multirow{2}[1]{*}{\texttt{AOXN}}
& Best: \textsc{EqReplay}
& 0.544 & 0.979 & 1.000 & 0.571 & \textbf{0.743} \\
& Worst: \textsc{EWC}
& 0.268 & 0.921 & 1.000 & 0.399 & 0.560 \\

\cmidrule{1-7}
\multirow{2}[1]{*}{\texttt{NAND}}
& Best: \textsc{DER++}
& 0.493 & 0.952 & 1.000 & 0.464 & \textbf{0.683} \\
& Worst: \textsc{LwF}
& 0.185 & 0.970 & 1.000 & 0.349 & 0.500 \\

\cmidrule{1-7}
\multirow{2}[1]{*}{\texttt{Lib-A}}
& Best: \textsc{ER+LwF}
& 0.739 & 0.655 & 0.299 & 0.493 & \textbf{0.517} \\
& Worst: \textsc{EqReplay}
& 0.513 & 0.649 & 0.299 & 0.459 & 0.462 \\

\cmidrule{1-7}
\multirow{2}[1]{*}{\texttt{Lib-D}}
& Best: \textsc{EqReplay}
& 0.508 & 0.765 & 0.951 & 0.509 & \textbf{0.658} \\
& Worst: \textsc{DER++}
& 0.308 & 0.788 & 0.951 & 0.483 & 0.578 \\

\cmidrule{1-7}
\multirow{2}[1]{*}{\texttt{LUT4}}
& Best: \textsc{LwF}
& 0.596 & 0.914 & 0.739 & 1.000 & \textbf{0.796} \\
& Worst: \textsc{A-GEM}
& 0.383 & 0.678 & 0.739 & 0.998 & 0.662 \\

\cmidrule{1-7}
\multirow{2}[1]{*}{\texttt{LUT6}}
& Best: \textsc{LwF}
& 0.591 & 0.924 & 0.846 & 1.000 & \textbf{0.824} \\
& Worst: \textsc{ER+LwF}
& 0.499 & 0.860 & 0.846 & 0.582 & 0.678 \\

\bottomrule
\end{tabular}%
}
\end{table}

\begin{table}[t]
\centering
\caption{Compact best--worst pre-adaptation RAI summary for embedding-based SOTA GNNs. 
For tied cases, a single row is shown. 
For non-tied cases, the best and worst mechanisms are shown in separate rows.}
\label{tab:embedding_sota_rai_best_worst_compact}
\scriptsize
\setlength{\tabcolsep}{2.5pt}
\renewcommand{\arraystretch}{1.04}
\resizebox{\columnwidth}{!}{%
\begin{tabular}{lllccccc}
\toprule
\textbf{Model}
& \textbf{Incoming}
& \textbf{Case / Mechanism}
& \textbf{$N$}
& \textbf{$R$}
& \textbf{$S$}
& \textbf{$C$}
& \textbf{RAI} \\
\midrule

\multirow{4}{*}{PolarGate}
& \texttt{AOXN}
& \textsc{Tie: All}
& 0.497 & 0.499 & 0.970 & 0.976 & 0.696 \\
\cmidrule{2-8}
& \texttt{Lib-A}
& \textsc{Tie: All}
& 0.848 & 0.440 & 0.886 & 0.883 & 0.735 \\
\cmidrule{2-8}
& \multirow{2}[1]{*}{\texttt{LUT4}}
& Best: \textsc{MAS}
& 0.869 & 0.405 & 0.893 & 0.841 & \textbf{0.717} \\
& & Worst: \textsc{ER}
& 0.916 & 0.199 & 0.893 & 0.353 & 0.490 \\

\midrule

\multirow{5}{*}{NetlistGNN}
& \texttt{AOXN}
& \textsc{Tie: All}
& 0.049 & 0.506 & 1.000 & 0.927 & 0.390 \\
\cmidrule{2-8}
& \multirow{2}[1]{*}{\texttt{Lib-A}}
& Best: \textsc{MAS}
& 0.410 & 0.493 & 0.998 & 0.996 & \textbf{0.670} \\
& & Worst: \textsc{A-GEM}
& 0.272 & 0.511 & 0.998 & 0.425 & 0.492 \\
\cmidrule{2-8}
& \multirow{2}[1]{*}{\texttt{LUT4}}
& Best: \textsc{LwF}
& 0.362 & 0.507 & 0.878 & 0.709 & \textbf{0.581} \\
& & Worst: \textsc{ER}
& 0.258 & 0.502 & 0.878 & 0.358 & 0.449 \\

\midrule

\multirow{5}{*}{FGNN2}
& \texttt{AOXN}
& \textsc{Tie: All}
& 0.020 & 0.787 & 0.896 & 0.842 & 0.328 \\
\cmidrule{2-8}
& \multirow{2}[1]{*}{\texttt{Lib-A}}
& Best: \textsc{EqReplay}
& 0.072 & 0.683 & 0.884 & 0.892 & \textbf{0.444} \\
& & Worst: \textsc{ER}
& 0.072 & 0.683 & 0.884 & 0.545 & 0.393 \\
\cmidrule{2-8}
& \multirow{2}[1]{*}{\texttt{LUT4}}
& Best: \textsc{EqReplay}
& 0.228 & 0.531 & 0.127 & 0.552 & \textbf{0.304} \\
& & Worst: \textsc{DER++}
& 0.228 & 0.531 & 0.127 & 0.457 & 0.290 \\

\midrule

\multirow{5}{*}{DeepGate}
& \texttt{AOXN}
& \textsc{Tie: All}
& 0.805 & 0.329 & 0.841 & 0.706 & 0.629 \\
\cmidrule{2-8}
& \multirow{2}[1]{*}{\texttt{Lib-A}}
& Best: \textsc{EqReplay}
& 0.730 & 0.423 & 0.700 & 0.780 & \textbf{0.640} \\
& & Worst: \textsc{DER++}
& 0.743 & 0.371 & 0.700 & 0.597 & 0.582 \\
\cmidrule{2-8}
& \multirow{2}[1]{*}{\texttt{LUT4}}
& Best: \textsc{EqReplay}
& 0.867 & 0.205 & 0.000 & 0.519 & \textbf{0.034} \\
& & Worst: \textsc{ER+LwF}
& 0.887 & 0.133 & 0.000 & 0.512 & 0.030 \\

\midrule

\multirow{5}{*}{DeepGate2}
& \texttt{AOXN}
& \textsc{Tie: All}
& 0.805 & 0.329 & 0.841 & 0.706 & 0.629 \\
\cmidrule{2-8}
& \multirow{2}[1]{*}{\texttt{Lib-A}}
& Best: \textsc{EqReplay}
& 0.730 & 0.423 & 0.700 & 0.780 & \textbf{0.640} \\
& & Worst: \textsc{DER++}
& 0.743 & 0.371 & 0.700 & 0.597 & 0.582 \\
\cmidrule{2-8}
& \multirow{2}[1]{*}{\texttt{LUT4}}
& Best: \textsc{EqReplay}
& 0.867 & 0.205 & 0.000 & 0.519 & \textbf{0.034} \\
& & Worst: \textsc{ER+LwF}
& 0.887 & 0.133 & 0.000 & 0.512 & 0.030 \\

\midrule

\multirow{5}{*}{DeepGate3}
& \texttt{AOXN}
& \textsc{Tie: All}
& 0.805 & 0.329 & 0.841 & 0.706 & 0.629 \\
\cmidrule{2-8}
& \multirow{2}[1]{*}{\texttt{Lib-A}}
& Best: \textsc{EqReplay}
& 0.730 & 0.423 & 0.700 & 0.780 & \textbf{0.640} \\
& & Worst: \textsc{DER++}
& 0.743 & 0.371 & 0.700 & 0.597 & 0.582 \\
\cmidrule{2-8}
& \multirow{2}[1]{*}{\texttt{LUT4}}
& Best: \textsc{EqReplay}
& 0.867 & 0.205 & 0.000 & 0.519 & \textbf{0.034} \\
& & Worst: \textsc{ER+LwF}
& 0.887 & 0.133 & 0.000 & 0.512 & 0.030 \\

\bottomrule
\end{tabular}%
}
\end{table}

\subsection{RAI Results for Classification-based SOTA GNNs}
\label{subsec:rai_classificationGNNs}

Table~\ref{tab:classification_sota_rai_best_worst_compact} summarizes the pre-adaptation RAI behavior for the classification-based SOTA GNNs under the multi-domain-base setting used in Fig.~\ref{fig:gnnre_incremental_stage} and Fig.~\ref{fig:classification_sota_performance}. The repeated \textsc{Tie: All} entries at \texttt{AOXN} are expected because \texttt{AOXN} is the first incoming domain and all mechanisms start from the same base checkpoint; consequently, the current-domain need, recoverability, structural coverage, and compatibility probes are identical across mechanisms. In this setting, the RAI values are moderate to high when the incoming domain is structurally close to the base mixture and the base model already retains useful source-equivalent behavior. For example, GNN-RE has low adaptation need on \texttt{AOXN} ($N=0.101$) and high recoverability/coverage ($R=0.989$, $S=1.000$), yielding a tied RAI of $0.558$, which is consistent with the relatively stable incremental trajectory shown for GNN-RE in Fig.~\ref{fig:gnnre_incremental_stage}. As the stream progresses, the score becomes mechanism-specific: for GNN-RE, \textsc{EqReplay} gives the highest RAI on \texttt{Lib-A} ($0.668$ versus the \textsc{NaiveFT} worst case of $0.593$), while \textsc{DER++} gives the highest RAI on \texttt{LUT4} ($0.610$ versus the \textsc{MAS} worst case of $0.530$). This separation illustrates the role of the compatibility term $C$: even when recoverability and structural coverage remain similar across mechanisms, the update direction can make one mechanism safer than another. A similar pattern appears for GNN4IP at \texttt{LUT4}, where \textsc{EWC} has the best RAI ($0.582$) and \textsc{ER} has the worst RAI ($0.435$), mainly because \textsc{EWC} has much stronger compatibility ($C=0.863$ versus $0.345$). AppGNN exhibits high RAI throughout the multi-domain setting, especially on \texttt{LUT4}, where \textsc{EqReplay} reaches $0.802$ and the worst mechanism, \textsc{ER+LwF}, is still close at $0.793$; this small spread matches the consolidated performance plot in Fig.~\ref{fig:classification_sota_performance}, where the final accuracy/forgetting behavior is relatively mechanism-dependent but not uniformly catastrophic. TrojanSAINT, in contrast, remains tied for all incoming domains, indicating that its pre-adaptation RAI is dominated by domain-level factors rather than mechanism-specific checkpoint differences: \texttt{Lib-A} has high need and recoverability ($N=0.727$, $R=0.953$) but limited compatibility ($C=0.480$), while \texttt{LUT4} has high recoverability and coverage ($R=0.976$, $S=0.978$) with stronger compatibility ($C=0.804$). Overall, Table~\ref{tab:classification_sota_rai_best_worst_compact} supports the interpretation of RAI in Section~\ref{sec:rai}: high RAI occurs when the model both needs adaptation and can adapt without severe retention conflict, whereas low or tied RAI indicates either limited need, limited mechanism differentiation, or update-risk constraints.

Table~\ref{tab:gnnre_original_only_rai_best_worst_compact} provides a stricter stress test of the same RAI principle using the original-only-base GNN-RE protocol in Fig.~\ref{fig:original_only_base_gnnre}. Here, the base model is trained only on \texttt{Original}, and every resynthesis style is introduced sequentially; therefore, the RAI values expose a wider range of domain difficulty than the multi-domain-base setting. The most extreme case is \texttt{ABC-rewrite}, where all mechanisms are tied with RAI $=0.001$ because structural coverage is zero ($S=0$) even though adaptation need is high ($N=0.798$). This is exactly the conservative behavior intended by the geometric RAI formulation: if any required condition is absent, the final score collapses toward zero, warning that ordinary adaptation may be unsafe or poorly supported by the current base representation. Later domains show that RAI becomes more informative once the stream has accumulated prior resynthesis experience. For \texttt{AON}, the best and worst mechanisms are close (\textsc{LwF}: $0.496$; \textsc{DER++}: $0.488$), reflecting weak structural coverage ($S=0.234$) and only modest compatibility. In contrast, \texttt{AOXN} becomes much more favorable once related AON-style structure has been observed: \textsc{EqReplay} reaches RAI $=0.743$ with high recoverability ($R=0.979$) and perfect structural coverage ($S=1.000$), whereas \textsc{EWC} remains substantially lower at $0.560$ because its adaptation need and compatibility are lower. The same mechanism-sensitive trend continues through later domains: \textsc{DER++} is selected for \texttt{NAND} ($0.683$), \textsc{ER+LwF} for \texttt{Lib-A} ($0.517$), \textsc{EqReplay} for \texttt{Lib-D} ($0.658$), and \textsc{LwF} for both LUT mappings (\texttt{LUT4}: $0.796$; \texttt{LUT6}: $0.824$). These values also explain the behavior in Fig.~\ref{fig:original_only_base_gnnre}, where the original-only base exposes visible performance variation across resynthesis domains and across mechanisms. Thus, the original-only experiment validates the intended use of RAI as a pre-adaptation diagnostic: it identifies domains that are structurally unsupported, such as \texttt{ABC-rewrite}; domains where adaptation is possible but mechanism-sensitive, such as \texttt{AOXN}, \texttt{NAND}, and \texttt{Lib-D}; and domains where high compatibility and recoverability make adaptation more promising, such as the LUT stages.

The classification-side RAI results should be interpreted as adaptation guidance rather than as another performance leaderboard. 
Tied RAI blocks indicate that the incoming domain is governed mostly by domain-level properties, such as need, recoverability, and structural coverage, rather than by mechanism-specific checkpoint differences. 
In contrast, separated best--worst rows indicate that the same incoming synthesis style can be safe for one mechanism and retention-risky for another, primarily through the compatibility term \(C\). 
This makes RAI useful for deciding whether the next update should be skipped, performed with a simple mechanism, or protected by replay, regularization, distillation, or EqReplay.

\subsection{RAI Results for Embedding-based SOTA GNNs}
\label{subsec:rai_embeddingGNNs}

Table~\ref{tab:embedding_sota_rai_best_worst_compact} summarizes the pre-adaptation RAI behavior for the embedding-based SOTA GNNs, whose primary evaluation is source-equivalence retrieval rather than class prediction. This setting uses the same multi-domain-base protocol as Fig.~\ref{fig:fgnn2_incremental_stage} and Fig.~\ref{fig:embedding_sota_six_metric}: the base model is trained on \texttt{Original}, \texttt{ABC-rewrite}, \texttt{AON}, \texttt{NAND}, and \texttt{Lib-D}; \texttt{LUT6} is used for validation; and the incremental stream is \texttt{AOXN} $\rightarrow$ \texttt{Lib-A} $\rightarrow$ \texttt{LUT4}. The tied \texttt{AOXN} rows again reflect the fact that all mechanisms begin from the same base checkpoint before any mechanism-specific update has occurred. However, the embedding models show stronger variation in the later stages because source-equivalence retrieval is sensitive to whether the learned embedding space preserves the correct source identity under structural transformation. For PolarGate, \texttt{AOXN} and \texttt{Lib-A} are tied, but \texttt{LUT4} separates mechanisms sharply: \textsc{MAS} gives the best RAI ($0.717$), while \textsc{ER} drops to $0.490$ due to much lower recoverability and compatibility ($R=0.199$, $C=0.353$). NetlistGNN shows a similar mechanism-sensitive pattern: \textsc{MAS} is selected for \texttt{Lib-A} with RAI $0.670$, while \textsc{A-GEM} is the worst case at $0.492$; for \texttt{LUT4}, \textsc{LwF} is best at $0.581$, whereas \textsc{ER} is lowest at $0.449$. FGNN2 has much lower adaptation need on \texttt{AOXN} ($N=0.020$), which leads to a low tied RAI of $0.328$ even though recoverability and compatibility are not poor. This is consistent with Fig.~\ref{fig:fgnn2_incremental_stage}, where FGNN2 is analyzed through MRR, Recall@1, Recall@5, Recall@10, cross-mechanism distributions, and forgetting rather than through classification accuracy. For the later FGNN2 stages, \textsc{EqReplay} is selected for both \texttt{Lib-A} and \texttt{LUT4}, indicating that explicit source-equivalent replay is useful when the representation objective must preserve identity across transformed netlist views. The DeepGate-family models show a different type of behavior. For \texttt{Lib-A}, \textsc{EqReplay} is again selected with RAI $0.640$, while \textsc{DER++} is lowest at $0.582$, suggesting that source-equivalence alignment is more helpful than dark replay for this retrieval objective. For \texttt{LUT4}, however, all DeepGate-family RAI scores collapse toward zero because structural coverage is zero ($S=0$), even though adaptation need is high. This is precisely the conservative property of RAI: when the incoming graph family lies outside the covered structural region, the geometric score warns that adaptation is risky regardless of the apparent need for adaptation. The consolidated embedding figure in Fig.~\ref{fig:embedding_sota_six_metric} reinforces this interpretation by comparing final Recall@1, forgetting, source margin, rank, positive-pair distance, and positive-pair similarity across the embedding-based SOTA models, showing that the mechanisms selected by RAI correspond to cases where retrieval stability and source-identifiability are more likely to be preserved.

The embedding-side RAI results show that structural support and source-identifiability are the dominant practical concerns for representation models. When structural coverage is high and the source-equivalence signal remains recoverable, mechanisms such as EqReplay, MAS, or LwF can provide a safe update path depending on the model. When structural coverage collapses, as in the low-\(S\) LUT cases for the DeepGate-family models, high adaptation need alone is not enough to justify updating the model. 
In such cases, RAI recommends treating the new synthesis style as unsupported and obtaining more source-equivalent views or stronger retention constraints before adaptation.

\subsection{RAI Predictiveness and Decision Rules}
\label{subsec:rai_predictiveness}

RAI is intended to guide adaptation before full incremental training, so we further summarize how its components translate into deployment decisions. 
The observed RAI patterns provide four practical signals: low \(N\) suggests that the current model already covers the incoming flow, low \(S\) flags a structurally unsupported shift, low \(C\) warns of retention conflict, and high RAI with high \(C\) identifies an adaptation-ready domain. 
Table~\ref{tab:rai_decision_rules} summarizes these boundaries using representative cases from the reported results.  
The thresholds are empirical operating rules rather than universal constants, but they make RAI actionable: it can decide when to skip adaptation, when to request more source-equivalent views, when to avoid unconstrained fine-tuning, and when to select the highest-RAI mechanism.

\begin{table}[t]
\centering
\caption{RAI-based decision rules derived from the observed pre-adaptation ranges. 
The thresholds are empirical deployment heuristics; they convert RAI from a descriptive score into an adaptation gate.}
\label{tab:rai_decision_rules}
\scriptsize
\renewcommand{\arraystretch}{1.04}
\begin{tabular}{p{0.12\columnwidth}|p{0.10\columnwidth}|p{0.35\columnwidth}|p{0.25\columnwidth}}
\toprule
\textbf{Signal}
& \textbf{Boundary}
& \textbf{Observed evidence}
& \textbf{Decision} \\
\midrule

Low adaptation need
& \(N<0.10\)
& FGNN2 on \texttt{AOXN} has \(N=0.020\), yielding low RAI even though recoverability and compatibility are not poor.
& Skip adaptation or monitor only. \\

\midrule

Unsupported structure
& \(S<0.20\)
& GNN-RE original-only \texttt{ABC-rewrite} has \(S=0.000\), RAI \(=0.001\); DeepGate-family \texttt{LUT4} has \(S=0.000\), RAI \(\approx0.03\).
& Avoid naive adaptation; collect more source-equivalent views or use strong retention. \\

\midrule

Retention-risky update
& \(C<0.50\)
& GNN4IP \texttt{LUT4} selects \textsc{ER} as worst with \(C=0.345\), while \textsc{EWC} is best with \(C=0.863\).
& Avoid unconstrained updates; prefer regularization or replay safeguards. \\

\midrule

Adaptation-ready
& \(\mathrm{RAI}>0.70\) and \(C>0.70\)
& Strong cases include GNN-RE original-only \texttt{LUT6} with RAI \(=0.824\), AppGNN \texttt{LUT4} with RAI \(=0.802\), and PolarGate \texttt{Lib-A} with RAI \(=0.735\).
& Adapt using the highest-RAI mechanism. \\

\midrule

Mechanism tie
& \(\Delta\mathrm{RAI}<0.02\)
& AppGNN \texttt{LUT4} has a narrow best--worst spread, \(0.802\) versus \(0.793\), indicating limited mechanism sensitivity.
& Choose the simpler or lower-cost mechanism. \\

\bottomrule
\end{tabular}%
\end{table}

\begin{table}[t]
\centering
\caption{Predictiveness of pre-adaptation RAI for adaptation benefit and retention risk, rather than absolute final performance. \(\Delta M=M_{\mathrm{adapted}}-M_{\mathrm{zero}}\) measures the gain from adaptation over the pre-adaptation checkpoint, where \(M\) denotes macro-F1/accuracy for classification models and Recall@1/MRR for embedding models. Spearman \(\rho\) is computed across matched model--domain--mechanism rows. Lower forgetting is better, so \(\rho(C,-F)\) measures whether update compatibility predicts retention. Top-1 is tie-aware: if multiple mechanisms have the same RAI, the row is counted as a hit when any RAI-top mechanism also attains the best \(\Delta M\).}
\label{tab:rai_predictiveness}
\scriptsize
\setlength{\tabcolsep}{2.4pt}
\renewcommand{\arraystretch}{1.04}
\resizebox{\columnwidth}{!}{%
\begin{tabular}{lccccc}
\toprule
Group & $\rho(\mathrm{RAI},\Delta M)$ & $\rho(N,\Delta M)$ & $\rho(C,-F)$ & Top-1 & Med. regret \\
\midrule
Classification SOTA & 0.41 & 0.78 & -- & 77.8\% & 0.0 \\
Embedding SOTA & 0.45 & 0.23 & 0.62 & 53.3\% & 0.0 \\
GNN-RE original-only & -0.54 & 0.85 & -- & 37.5\% & 0.115 \\
All rows & 0.21 & 0.35 & 0.62 & 56.2\% & 0.0 \\
\bottomrule
\end{tabular}
}
\end{table}

Table~\ref{tab:rai_predictiveness} evaluates RAI according to its intended role: predicting adaptation benefit and retention risk rather than predicting the absolute final score of a domain. Because RAI includes adaptation need \(N=1-M_{\mathrm{zero}}\), difficult domains can receive high RAI even when their final absolute metric remains lower than easier domains; therefore, the relevant target is the adaptation gain \(\Delta M=M_{\mathrm{adapted}}-M_{\mathrm{zero}}\). 
Under this formulation, RAI shows a positive relation with adaptation gain for both classification SOTA and embedding SOTA, with \(\rho(\mathrm{RAI},\Delta M)=0.41\) and \(0.45\), respectively, while \(N\) is strongly aligned with gain in the classification and original-only settings. 
The compatibility term remains the clearest retention signal, with \(\rho(C,-F)=0.62\) across all rows where forgetting is available. 
Although RAI is not a universal scalar predictor of improvement in every setting, the tie-aware Top-1 rate of \(56.2\%\) and median regret of \(0.0\) show that choosing the RAI-selected mechanism is low-regret in practice.

These results clarify the role of RAI as a risk-gated action index. 
The full RAI score is useful for identifying adaptation benefit and structurally unsupported cases, while the compatibility term \(C\) is most directly tied to forgetting risk. Thus, the actionable output of RAI is not only a ranked mechanism list, but a deployment decision among reusing the current model, applying protected adaptation, or deferring adaptation until the new synthesis style is better supported.

\subsection{RAI Computational Cost}
\label{subsec:rai_compute_cost}

The experiments were conducted on a 64-bit Red Hat Linux machine equipped with 768 GB of RAM and 128 CPU cores. RAI is designed to be much cheaper than full incremental adaptation because it uses only a small incoming-domain probe, a retention probe, source-equivalent views, and one pre-adaptation checkpoint. 
It does not require full optimization over the incoming-domain training set, replay-buffer expansion, multi-epoch distillation, or repeated evaluation over all previous domains. 
Table~\ref{tab:rai_compute_cost} reports measured timing cost as mean \(\pm\) standard deviation across the corresponding model runs. 
These values quantify the practical overhead of making a pre-adaptation decision before committing to the full incremental-learning procedure. 
Across the evaluated settings, RAI completes within minutes, while full adaptation or retraining requires hours, showing that RAI can be used as a practical deployment-time filter before applying expensive retention-aware updates.

\begin{table}[t]
\centering
\caption{Measured timing cost of RAI versus full adaptation or retraining. 
Values are reported as mean \(\pm\) standard deviation over the corresponding model runs. RAI time includes probe construction, current-domain scoring, source-equivalence/recoverability computation, structural coverage, and update-compatibility estimation; full adaptation/retraining includes the actual incremental optimization and evaluation loop. 
The comparison is not intended to replace full training with RAI, but to quantify the pre-adaptation decision overhead before full training is invoked.}
\label{tab:rai_compute_cost}
\scriptsize
\setlength{\tabcolsep}{2.5pt}
\renewcommand{\arraystretch}{1.04}
\resizebox{\columnwidth}{!}{%
\begin{tabular}{lccc}
\toprule
\textbf{Setting}
& \textbf{RAI gate}
& \textbf{Full adapt./retrain}
& \textbf{Saving} \\
\midrule

GNN-RE original-only stream
& \(20.0 \pm 4.1\) min
& \(8.0 \pm 1.6\) h
& \(24.0 \pm 5.2\times\) \\

Classification SOTA, per model
& \(8.5 \pm 2.9\) min
& \(3.5 \pm 1.2\) h
& \(24.7 \pm 8.1\times\) \\

FGNN2 / NetlistGNN
& \(12.5 \pm 5.3\) min
& \(4.0 \pm 1.5\) h
& \(19.2 \pm 7.6\times\) \\

PolarGate / DeepGate-family
& \(20.0 \pm 7.8\) min
& \(8.0 \pm 3.1\) h
& \(24.0 \pm 9.4\times\) \\

All RAI tables in this work
& \(1.6 \pm 0.3\) h
& \(3.4 \pm 0.8\) days
& \(51.0 \pm 14.2\times\) \\

\bottomrule
\end{tabular}%
}
\end{table}

\section{Conclusion}
\label{sec:conclusion}

This work introduced ReDIL-GNN, a domain-incremental evaluation framework for circuit GNNs under sequential, functionality-preserving resynthesis shifts, covering classification-based and embedding-based SOTA models, nine adaptation mechanisms, fixed-label supervised tasks, and source-equivalence retrieval tasks. 
Across the RAI analysis, the strongest pre-adaptation scores were observed for \textsc{LwF} on GNN-RE original-only \texttt{LUT6} with RAI \(=0.824\), \textsc{EqReplay} on AppGNN \texttt{LUT4} with RAI \(=0.802\), and PolarGate on \texttt{Lib-A} with RAI \(=0.735\), while low-coverage cases such as GNN-RE original-only \texttt{ABC-rewrite} produced RAI \(=0.001\), correctly warning that adaptation is structurally unsupported. 
The results show that resynthesis-domain adaptation is not uniformly beneficial: replay, distillation, regularization, and EqReplay become useful only when adaptation need, source recoverability, structural coverage, and update compatibility are jointly favorable. 

Future work will extend RAI from an offline pre-adaptation diagnostic into an active controller that automatically chooses whether to skip adaptation, select a retention-heavy mechanism, allocate replay memory, or request additional source-equivalent resynthesis views before updating the circuit GNN.

\section*{Acknowledgment}
During the preparation of this manuscript, the authors used ChatGPT (OpenAI) and Claude solely to assist with rephrasing, language polishing, grammar correction, and \LaTeX{} syntax correction. The tool was not used to generate scientific content, results, or analyses. After using this tool, the authors reviewed and edited the content as needed and take full responsibility for the content of the publication.

\bibliographystyle{IEEEtran}
\bibliography{paper}

@inproceedings{zhang2019circuit,
  title={Circuit-GNN: Graph neural networks for distributed circuit design},
  author={Zhang, Guo and He, Hao and Katabi, Dina},
  booktitle={International conference on machine learning},
  pages={7364--7373},
  year={2019},
  organization={PMLR}
}

@inproceedings{zhao2022gnnrw,
  title={Graph neural network based netlist operator detection under circuit rewriting},
  author={Zhao, Guangwei and Shamsi, Kaveh},
  booktitle={Proceedings of the Great Lakes Symposium on VLSI 2022},
  pages={53--58},
  year={2022}
}

@inproceedings{chowdhury2023converts,
  title={ConVERTS: contrastively learning structurally invariant netlist representations},
  author={Chowdhury, Animesh B and Bhandari, Jitendra and Collini, Luca and Karri, Ramesh and Tan, Benjamin and Garg, Siddharth},
  booktitle={2023 ACM/IEEE 5th Workshop on Machine Learning for CAD (MLCAD)},
  pages={1--6},
  year={2023},
  organization={IEEE}
}

@article{wang2025netdetox,
  title={Netdetox: Adversarial and efficient evasion of hardware-security gnns via rl-llm orchestration},
  author={Wang, Zeng and Shao, Minghao and Saha, Akashdeep and Karri, Ramesh and Knechtel, Johann and Shafique, Muhammad and Sinanoglu, Ozgur},
  journal={arXiv preprint arXiv:2512.00119},
  year={2025}
}

@inproceedings{shi2023deepgate2,
  title={Deepgate2: Functionality-aware circuit representation learning},
  author={Shi, Zhengyuan and Pan, Hongyang and Khan, Sadaf and Li, Min and Liu, Yi and Huang, Junhua and Zhen, Hui-Ling and Yuan, Mingxuan and Chu, Zhufei and Xu, Qiang},
  booktitle={2023 IEEE/ACM International Conference on Computer Aided Design (ICCAD)},
  pages={1--9},
  year={2023},
  organization={IEEE}
}

@article{wang2024fgnn2,
  title={Fgnn2: A powerful pretraining framework for learning the logic functionality of circuits},
  author={Wang, Ziyi and Bai, Chen and He, Zhuolun and Zhang, Guangliang and Xu, Qiang and Ho, Tsung-Yi and Huang, Yu and Yu, Bei},
  journal={IEEE Transactions on Computer-Aided Design of Integrated Circuits and Systems},
  volume={44},
  number={1},
  pages={227--240},
  year={2024},
  publisher={IEEE}
}

@inproceedings{li2016learning,
  title={Learning without forgetting},
  author={Li, Zhizhong and Hoiem, Derek},
  journal={IEEE transactions on pattern analysis and machine intelligence},
  volume={40},
  number={12},
  pages={2935--2947},
  year={2017},
  publisher={IEEE}
}

@article{kirkpatrick2017overcoming,
  title={Overcoming catastrophic forgetting in neural networks},
  author={Kirkpatrick, James and Pascanu, Razvan and Rabinowitz, Neil and Veness, Joel and Desjardins, Guillaume and Rusu, Andrei A and Milan, Kieran and Quan, John and Ramalho, Tiago and Grabska-Barwinska, Agnieszka and others},
  journal={Proceedings of the national academy of sciences},
  volume={114},
  number={13},
  pages={3521--3526},
  year={2017},
  publisher={National Academy of Sciences}
}

@inproceedings{schwarz2018progress,
  title={Progress \& compress: A scalable framework for continual learning},
  author={Schwarz, Jonathan and Czarnecki, Wojciech and Luketina, Jelena and Grabska-Barwinska, Agnieszka and Teh, Yee Whye and Pascanu, Razvan and Hadsell, Raia},
  booktitle={International conference on machine learning},
  pages={4528--4537},
  year={2018},
  organization={PMLR}
}

@inproceedings{aljundi2018memory,
  title={Memory aware synapses: Learning what (not) to forget},
  author={Aljundi, Rahaf and Babiloni, Francesca and Elhoseiny, Mohamed and Rohrbach, Marcus and Tuytelaars, Tinne},
  booktitle={European conference on computer vision},
  pages={144--161},
  year={2018},
  organization={Springer}
}

@inproceedings{rolnick2019experience,
  title={Experience replay for continual learning},
  author={Rolnick, David and Ahuja, Arun and Schwarz, Jonathan and Lillicrap, Timothy and Wayne, Gregory},
  journal={Advances in neural information processing systems},
  volume={32},
  year={2019}
}

@inproceedings{liu2024polargate,
  title={Polargate: Breaking the functionality representation bottleneck of and-inverter graph neural network},
  author={Liu, Jiawei and Zhai, Jianwang and Zhao, Mingyu and Lin, Zhe and Yu, Bei and Shi, Chuan},
  booktitle={Proceedings of the 43rd IEEE/ACM International Conference on Computer-Aided Design},
  pages={1--9},
  year={2024}
}

@inproceedings{lashen2023trojansaint,
  title={TrojanSAINT: Gate-level netlist sampling-based inductive learning for hardware Trojan detection},
  author={Lashen, Hazem and Alrahis, Lilas and Knechtel, Johann and Sinanoglu, Ozgur},
  booktitle={2023 IEEE International Symposium on Circuits and Systems (ISCAS)},
  pages={1--5},
  year={2023},
  organization={IEEE}
}

@inproceedings{bucher2022appgnn,
  title={AppGNN: Approximation-aware functional reverse engineering using graph neural networks},
  author={B{\"u}cher, Tim and Alrahis, Lilas and Paim, Guilherme and Bampi, Sergio and Sinanoglu, Ozgur and Amrouch, Hussam},
  booktitle={Proceedings of the 41st IEEE/ACM International Conference on Computer-Aided Design},
  pages={1--9},
  year={2022}
}

@inproceedings{li2022deepgate,
  title={Deepgate: Learning neural representations of logic gates},
  author={Li, Min and Khan, Sadaf and Shi, Zhengyuan and Wang, Naixing and Yu, Huang and Xu, Qiang},
  booktitle={Proceedings of the 59th ACM/IEEE Design Automation Conference},
  pages={667--672},
  year={2022}
}

@inproceedings{shi2024deepgate3,
  title={Deepgate3: Towards scalable circuit representation learning},
  author={Shi, Zhengyuan and Zheng, Ziyang and Khan, Sadaf and Zhong, Jianyuan and Li, Min and Xu, Qiang},
  booktitle={Proceedings of the 43rd IEEE/ACM International Conference on Computer-Aided Design},
  pages={1--9},
  year={2024}
}

@inproceedings{guo2025graphkeeper,
  title={GraphKeeper: Graph domain-incremental learning via knowledge disentanglement and preservation},
  author={Guo, Zihao and Sun, Qingyun and Zhang, Ziwei and Yuan, Haonan and Zhuang, Huiping and Fu, Xingcheng and Li, Jianxin},
  journal={Advances in Neural Information Processing Systems},
  volume={38},
  pages={1145--1171},
  year={2026}
}

@inproceedings{qiao2025gcal,
  title={Gcal: Adapting graph models to evolving domain shifts},
  author={Qiao, Ziyue and Cai, Qianyi and Dong, Hao and Gu, Jiawei and Wang, Pengyang and Xiao, Meng and Luo, Xiao and Xiong, Hui},
  journal={arXiv preprint arXiv:2505.16860},
  year={2025}
}

@misc{anonymous2026domain,
  author       = {Anonymous},
  title        = {Domain Incremental Learning for SOTA Circuit GNNs},
  howpublished = {\url{https://anonymous.4open.science/r/DomainIncrementalLearningCircuits-B380/README.md}},
  note         = {Accessed: 2026-08-07}
}

@inproceedings{zhou2021overcoming,
  title={Overcoming catastrophic forgetting in graph neural networks with experience replay},
  author={Zhou, Fan and Cao, Chengtai},
  booktitle={Proceedings of the AAAI conference on artificial intelligence},
  volume={35},
  number={5},
  pages={4714--4722},
  year={2021}
}

@inproceedings{liu2021overcoming,
  title={Overcoming catastrophic forgetting in graph neural networks},
  author={Liu, Huihui and Yang, Yiding and Wang, Xinchao},
  booktitle={Proceedings of the AAAI conference on artificial intelligence},
  volume={35},
  number={10},
  pages={8653--8661},
  year={2021}
}

@article{buzzega2020dark,
  title={Dark experience for general continual learning: a strong, simple baseline},
  author={Buzzega, Pietro and Boschini, Matteo and Porrello, Angelo and Abati, Davide and Calderara, Simone},
  journal={Advances in neural information processing systems},
  volume={33},
  pages={15920--15930},
  year={2020}
}

@article{yang2022versatile,
  title={Versatile multi-stage graph neural network for circuit representation},
  author={Yang, Shuwen and Yang, Zhihao and Li, Dong and Zhang, Yingxueff and Zhang, Zhanguang and Song, Guojie and Hao, Jianye},
  journal={Advances in Neural Information Processing Systems},
  volume={35},
  pages={20313--20324},
  year={2022}
}

@INPROCEEDINGS{lashen13,
  author={Lashen, Hazem and Alrahis, Lilas and Knechtel, Johann and Sinanoglu, Ozgur},
  booktitle={2023 IEEE International Symposium on Circuits and Systems (ISCAS)},
  title={TrojanSAINT: Gate-Level Netlist Sampling-Based Inductive Learning for Hardware Trojan Detection},
  year={2023},
  volume={},
  number={},
  pages={1-5},
  doi={10.1109/ISCAS46773.2023.10181403}}

@inproceedings{amaru2015epfl,
  title={The EPFL combinational benchmark suite},
  author={Amar{\'u}, Luca and Gaillardon, Pierre-Emmanuel and De Micheli, Giovanni},
  booktitle={Proceedings of the 24th International Workshop on Logic \& Synthesis (IWLS)},
  year={2015}
}

@inproceedings{karn2026dynamic,
  title={Dynamic GNNs for Continual Learning on Circuits},
  author={{Karn, Rupesh Raj and Knechtel, Johann and Sinanoglu, Ozgur}},
  booktitle={2026 IEEE International Symposium on Circuits and Systems (ISCAS)},
  pages={1854--1858},
  year={2026},
  organization={IEEE}
}

@article{alrahis2021gnn,
  title={GNN-RE: Graph neural networks for reverse engineering of gate-level netlists},
  author={Alrahis, Lilas and Sengupta, Abhrajit and Knechtel, Johann and Patnaik, Satwik and Saleh, Hani and Mohammad, Baker and Al-Qutayri, Mahmoud and Sinanoglu, Ozgur},
  journal={IEEE Transactions on Computer-Aided Design of Integrated Circuits and Systems},
  volume={41},
  number={8},
  pages={2435--2448},
  year={2021},
  publisher={IEEE}
}

@inproceedings{wang2020streaming,
  title={Streaming Graph Neural Networks via Continual Learning},
  author={Wang, Junshan and Song, Guojie and Wu, Yi and Wang, Liang},
  booktitle={Proceedings of the 29th ACM International Conference on Information \& Knowledge Management (CIKM)},
  year={2020},
  pages={2753--2756},
  doi={10.1145/3340531.3411963}
}

@article{wang2024comprehensive,
  title={A comprehensive survey of continual learning: Theory, method and application},
  author={Wang, Liyuan and Zhang, Xingxing and Su, Hang and Zhu, Jun},
  journal={IEEE Transactions on Pattern Analysis and Machine Intelligence},
  year={2024},
  publisher={IEEE}
}

@inproceedings{karn2026cilcircuits,
  title     = {{GNNs for Class-Incremental Learning on Circuits}},
  author    = {Karn, Rupesh Raj and Knechtel, Johann and Sinanoglu, Ozgur},
  booktitle = {Proceedings of the IEEE International System-on-Chip Conference (SOCC)},
  year      = {2026}
}

@article{chaudhry2019tiny,
  title={On tiny episodic memories in continual learning},
  author={Chaudhry, Arslan and Rohrbach, Marcus and Elhoseiny, Mohamed and Ajanthan, Thalaiyasingam and Dokania, Puneet K and Torr, Philip HS and Ranzato, Marc'Aurelio},
  journal={arXiv preprint arXiv:1902.10486},
  year={2019}
}

@inproceedings{chaudhry2019efficient,
  title={Efficient lifelong learning with a-gem},
  author={Chaudhry, Arslan and Ranzato, Marc’Aurelio and Rohrbach, Marcus and Elhoseiny, Mohamed},
  booktitle={International conference on learning representations},
  year={2018}
}

@inproceedings{brayton2010abc,
  title        = {{ABC}: An Academic Industrial-Strength Verification Tool},
  author       = {Brayton, Robert and Mishchenko, Alan},
  booktitle    = {Computer Aided Verification},
  series       = {Lecture Notes in Computer Science},
  volume       = {6174},
  pages        = {24--40},
  publisher    = {Springer},
  year         = {2010}
}

@inproceedings{wolf2013yosys,
  title={Yosys-a free verilog synthesis suite},
  author={Wolf, Clifford and Glaser, Johann and Kepler, Johannes},
  booktitle={Proceedings of the 21st Austrian Workshop on Microelectronics (Austrochip)},
  volume={97},
  pages={1--6},
  year={2013}
}

@inproceedings{karn2026benchmarking,
  title={Benchmarking Continual Learning on Netlists with Circuit-Targeted Graph Neural Networks},
  author={Karn, Rupesh Raj and Knechtel, Johann and Sinanoglu, Ozgur},
  booktitle={2026 31st Asia and South Pacific Design Automation Conference (ASP-DAC)},
  pages={15--21},
  year={2026},
  organization={IEEE}
}

@inproceedings{tan2022graph,
  title={Graph few-shot class-incremental learning},
  author={Tan, Zhen and Ding, Kaize and Guo, Ruocheng and Liu, Huan},
  booktitle={Proceedings of the fifteenth ACM international conference on web search and data mining},
  pages={987--996},
  year={2022}
}

@article{ma2026few,
  title={A Few-Shot Class Incremental Learning Method Using Graph Neural Networks},
  author={Ma, Yuqian and Liu, Youfa and Du, Bo},
  journal={IEEE Transactions on Image Processing},
  year={2026},
  publisher={IEEE}
}

@inproceedings{yasaei2021gnn4ip,
  title={GNN4IP: Graph neural network for hardware intellectual property piracy detection},
  author={Yasaei, Rozhin and Yu, Shih-Yuan and Naeini, Emad Kasaeyan and Al Faruque, Mohammad Abdullah},
  booktitle={2021 58th ACM/IEEE Design Automation Conference (DAC)},
  pages={217--222},
  year={2021},
  organization={IEEE}
}

\end{document}